# Enhanced Emotion Enabled Cognitive Agent Based Rear End Collision Avoidance Controller for Autonomous Vehicles

Faisal Riaz[1], Muaz A. Niazi[2]

[1]Dept. Of Computing-Iqra University, Islamabad, Pakistan

[2]Dept. Of Computer Sciences-COMSATS, Islamabad, Pakistan

Email: [1]fazi_ajku@yahoo.com, *muaz.niazi@gmail.com

**Abstract**

Rear end collisions are deadliest in nature and cause most of traffic casualties and injuries. In the existing research, many rear end collision avoidance solutions have been proposed. However, the problem with these proposed solutions is that they are highly dependent on precise mathematical models. Whereas, the real road driving is influenced by non-linear factors such as road surface situations, driver reaction time, pedestrian flow and vehicle dynamics, hence obtaining the accurate mathematical model of the vehicle control system is challenging. This problem with precise control based rear end collision avoidance schemes has been addressed using fuzzy logic, but the excessive number of fuzzy rules straightforwardly prejudice their efficiency. Furthermore, these fuzzy logic based controllers have been proposed without using proper agent based modeling that helps in mimicking the functions of an artificial human driver executing these fuzzy rules. Keeping in view these limitations, we have proposed an Enhanced Emotion Enabled Cognitive Agent (EEEC_Agent) based controller that helps the Autonomous Vehicles (AVs) to perform rear end collision avoidance with less number of rules, designed after fear emotion, and high efficiency. To introduce a fear emotion generation mechanism in EEEC_Agent, Orton, Clore & Collins (OCC) model has been employed. The fear generation mechanism of EEEC_Agent has been verified using NetLogo simulation. Furthermore, practical validation of EEEC_Agent functions has been performed using specially built prototype AV platform. Eventually, the qualitative comparative study with existing state of the art research works reflect that proposed model outperforms recent research.

## 1. Introduction

Rear end collisions are deadliest in nature and cause most of traffic casualties and injuries. According to Meng et al. (1), rear-end crashes constitute around 70% out of all the crashes happened in the tunnels – crossings and are the main cause of deaths and injuries. In another research work, Harb et al. (2) noted that rear end collisions alone contributed one-third of the 6 million stated crashes in the USA in the year of 2003. In addition, Chen et al. (3) reported that rear end collisions cause 1.078 million injuries in the USA only. Whereas, according to Nishimura et al. (4), front-rear end collisions have a substantial contribution in automotive-related trauma and long term injuries than other types of road collisions. From these statistics, it is obvious that how important is to tailor the efficient rear end collision avoidance solutions.

In the existing literature, many rear end collision avoidance solutions have been proposed. Moon et al. (5), have proposed proportional–integral–derivative based rear end collision avoidance controller that helps in avoiding the vehicles moving within Adaptive Cruise Control (ACC) mode. Gracia et al. (6), have proposed sliding mode control based rear end collision avoidance solution. In other research work, Van et al. (7) have proposed rear end collision avoidance between vehicles using linear quadratic optimal control technique. However, the problem with these proposed rear end collision avoidance solutions is that they are highly dependent on precise mathematical models (8). Whereas, the real road driving is influenced by non-linear factors like road surface situations, driver reaction time, pedestrian flow and vehicle dynamics, hence obtaining the accurate mathematical model of the vehicle control system is difficult.

This problem with precise control based rear end collision avoidance schemes has been addressed using fuzzy logic due to its capability of handling nonlinear systems and capturing the driving characteristics. Milanes et al. (9), have proposed rear end collision detection and avoidance system using fuzzy logic and keeping in view the human driver behavior during rear-end collision avoidance. Toshihisa and Motoyuki (10) has modelled the human driver characteristics like driving style, reaction time and cognitive state using fuzzy logic to propose the rear end collision avoidance scheme. In the same way, Razzaq et al. (11) have proposed agent based model of rear end collision avoidance by building fuzzy rules considering the human driver factors along the different weather conditions. Though the fuzzy logic based rear end collision avoidance schemes resolve the issue of mathematical based rear end collision avoidance schemes but still suffer from critical problems.

The first main problem is that the fuzzy logic based rear end collision avoidance controllers rely on the number of fuzzy rules, an excessive number of such will straightforwardly prejudice its efficiency. Furthermore, with the increase in the fuzzy subsets, the fuzzy rules increases exponentially, which increases the computation time of rear end collision avoidance decision making. Whereas, in the real-time environment of road traffic, the decision making should be very quick to avoid the collisions and so the fuzzy logic based rear end collisions avoidance schemes failed to be a suitable candidate for this purpose. The second main problem is that though the fuzzy logic based controllers, fuzzy rules, have been modelled after the human drivers' behavior while ignoring the impact of emotions in the decision making of human drivers. The third main problem is that the existing FLCs have been designed for semi-autonomous vehicles and a complete human independent, agent based, and real time rear end collision avoidance controller for fully AVs, as making roads safer by avoiding road collisions is one of the main motivations for inventing Autonomous Vehicles (AVs) (12), has not been proposed and validated at theoretical and practical level, which act as a basis for industry level development of AVs. Hence there is a need to have a novel human emotions inspired rear end collision avoidance scheme for fully AVs, which overcomes the above-mentioned problems of both mathematical and fuzzy logic based rear end collision avoidance schemes.

However, a question arises that humans are the weakest element in driving and why it is useful to produce a collision avoidance component that acts like a human. No doubt that the Human drivers are the main cause of road accidents because of many reasons like texting during driving (13), paying attention towards bill boards (14), and mobile phone conversation (15), but still the expertise and flexible models of humans can be classified as state of the art for designing the different components of autonomous systems and most of the latest work in the field of AVs like (9) - (12) and (16) - (17) are considering this aspect. Furthermore, another question arises that automated driving works because the human weaknesses like too long time to reason or emotions that produce irrational, slow reactions are out of the play. The answer is that we have replaced the human drivers with a computer-based system, which emulate the role of human emotions in making robust rear end collision avoidance decisions. In other words, we are exploring the role of irrationality in making rational decisions and here rational decisions refer the rear end collision decisions.

**Motivation behind enhancing existing emotions enabled agents:** The notion of exploring the emotions for traffic and other real world problems is not a new one. Brain Emotional Learning (BEL) has been utilised by Shahmirzadi and Langari in (17) to propose intelligent sliding mode control for rollover prevention in tractor-semitrailers. Though the authors have utilised the human brain-inspired emotion generation system, but have not utilised any well-investigated emotion computational mechanism, which helps in generating the emotions according to the rapid changes in the operating environment of vehicles. Furthermore, emotions eliciting and intensity computation mechanism has not been provided. In Rizzi et al. (18), have proposed situation appraisal based fear generation to control the motion of their robot. The proposed situation appraisal theory is based on a mathematical model. Whereas, in practical, the factors that contribute to emotion generation include different events, interaction patterns with entities and altruistic considerations. Because most of these factors are non-linear and time varying, a mathematical model cannot cover all of these aspects. Hence there is a strong need for some existing well investigated emotion appraisal theory to propose the authentic emotion generation mechanism. Furthermore, Rizzi et al. (18), have not proposed emotions eliciting and intensity computation mechanism, which help in computing different intensity levels of emotions. Leu et al. (19), have modelled artificial driver with emotions and personality. The purpose of the research work is to study the behavioural aspects of human drivers and how their collective behaviour affects traffic performance. For this purpose, they have proposed a cognitive-affective inspired Driver mental model. Though, the authors have proposed a mental model of human drivers but failed to model the real emotion generation mechanism, as humans have the inbuilt cognitive structure of emotions, which help them to generate emotions according to the varying environment and its entities. Also, emotion-eliciting and emotions intensity computation mechanism has not been proposed. In other research work, Riaz et al. (20) have introduced the emotions by the cognitive agent for efficient lateral collision avoidance between semi-AVs. The proposed Enhanced Emotion Enable Cognitive Agent (EEC_Agent) is very much inspired from BEL model with architectural features. Though the authors claimed the use of emotions in their agent, but no formal model of emotions has been presented. Furthermore, emotion-eliciting and emotion intensity computation mechanism has not been presented as well.

**Contribution:** Keeping in view the above-mentioned limitations of different research works related to the rear end collision avoidance and road safety techniques, we have made the following contributions to propose an efficient, practical and validated rear end collision avoidance solution for AVs.

- The Agent Architecture of human brain-inspired EEEC_Agent for rear end collision avoidance has been proposed.
- A proper fear emotion elicitation and generation mechanism, for EEC_Agent, has been proposed using OCC model (21).
- A proper Quantitative Computation mechanism to compute the different intensity levels of fear emotion has been proposed using fuzzy logic.
- A set of less number of rear-end collision avoidance rules, as compared to (8) and (9), influenced by fear intensity levels has been proposed.

- The verification of emotion generation mechanism of EEEC_Agent using NetLogo simulation has been performed.
- Proposed EEEC_Agent architecture inspired AV prototype has been built.
- The rigour validation of fear generation mechanism of EEEC_Agent, proposed rear end collision avoidance rules and tweak handling capability of EEEC_Agent in the sudden appearance of the pedestrian has been performed in a real road environment using prototype AV.
- The qualitative comparative study of proposed enhanced EEEC_Agent with existing state of the art emotions based agents and rear end collision avoidance schemes.

The rest of the paper has been organized as follows. Section 2 discusses proposed research method. Section 3 presents proposed agent-based model of emotions inspired collision avoidance controller. Section 4 presents experiments. Section 5 discusses the results and discussions. Practical validation of the proposed EEEC_Agent controller has been performed in section 6. General discussion in comparison with the different state of the art research works has been elucidated in section 7. The paper concludes in Section 8.

## 2. Proposed Research Method

The proposed method has been presented in the figure 1. First of all detailed literature review has been performed to note the limitations of existing rear end collision avoidance solutions for semi and full AVs. In addition, we have performed a detailed literature review of emotions enabled agents, proposed for traffic and other problems, along with their limitations. We have employed exploratory agent-based modelling, CABC framework (22), to propose the architecture of EEEC_Agent based rear end collision avoidance controller. To overcome the limitations, find out with the help of both literature reviews, following enhancements have been made in the emotions enabled agent. First of all emotion generation mechanism has been introduced using appraisal theory and for this purpose, we have utilised OCC model (21). In next step, emotion intensity computation and emotion elicitation mechanisms have been devised and for this purpose, we have utilised fuzzy logic. Furthermore, to act efficiently with less computational time fear intensity level inspired rear end collision avoidance rules have been proposed as well. Then to verify the fear generation mechanism of emotion enabled agent NetLogo simulation has been designed. To provide the results of fear intensity levels, to the NetLogo simulation, which have been computed using fuzzy logic, the SimConnect approach has been utilised. Furthermore, to validate the proposed emotions inspired controller, we built a prototype AV platform. Then using prototype AV, we verified the performance of proposed rear end collision avoidance controller. In the last, the comparative study of the enhancements made in rear-end collision avoidance agent has been performed with existing state of the art research work.

## 3. Proposed Agent-Based Model of Emotions Inspired Collision Avoidance Controller

In this section, the agent-based model of EEEC_Agent has been proposed. The proposed EEEC_Agent is envisaged as an enhanced version of emotions inspired cognitive agent to avoid rear end road accidents. The proposed EEEC_Agent possesses human-like ability to feel the fear but act better than human.

### 3.1 Preliminary Discussion

The Neuro-imaging literature indicates that an almond-shaped structure in the human brain known as Amygdala is responsible for emotions processing. Different research studies (23) (24) show that Amygdala plays a vital role in decision making, memory processing and emotional reaction to the internally or externally generated stimuli. When human beings perceive danger via some input stimulus, the information is sent to the human brain via two paths commonly known as the Primary (or short) and Secondary (or long) path shown. The information conveyed via the short path activates the emotional response before an actual picture of the danger is made in the mind.

In primary route information is directly passed by sensory thalamus to Amygdala from where the action against stimuli is taken. Whereas in the long route of information processing, information is passed from sensory thalamus to the primary sensory cortex, uni-modal association cortex, poly-modal association cortex, then hippocampus and finally to Amygdala from where final action is passed to motor areas of the brain (23). It is important to mention that primary route is faster than a secondary route. Inspired by the fast emotion generation architecture of the human brain, basic EEC_Agent was devised by Shahmirzadi and Langari in (17) (20). However, the proper emotion generation mechanism along with emotion-eliciting and emotions, the intensity computation mechanism has been missed.

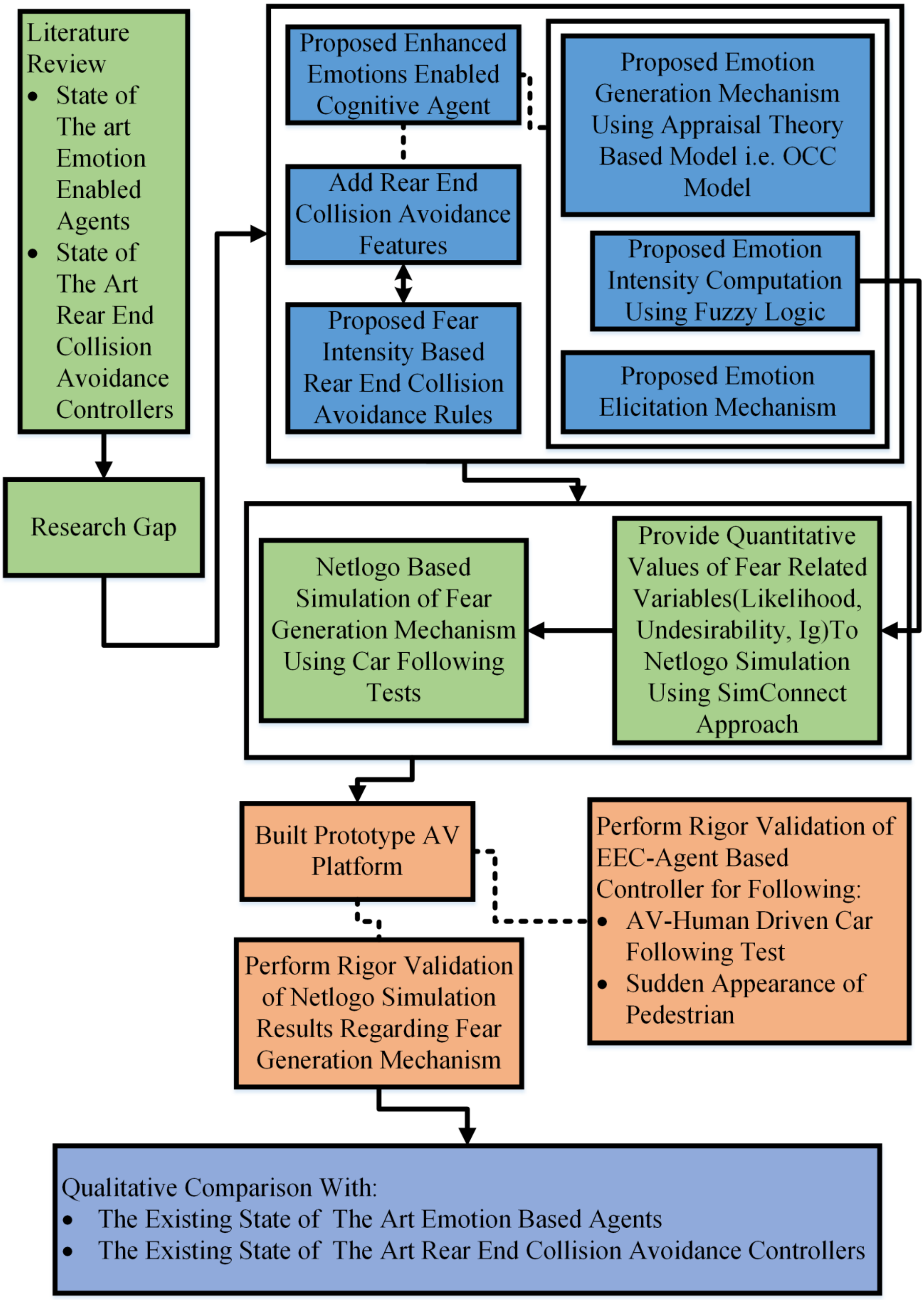

Figure. 1 Proposed Research Method

### 3.2 EEEC Agent-Based Rear End Collision Avoidance Controller Architecture

The architecture of the EEEC_Agent based rear end collision avoidance controller has been presented in figure 2. For the sake of simplicity, only two main neurones are considered, i.e. Hypothalamus neurone (NH) and Amygdala neurone (NA), because of their direct relation to the short route functionality of human brain in emergencies. In addition to generating the notion of the OCC model (21) defined prospect based emotion, i.e. Fear, in an EEEC_Agent, the methodology of Intercalated (ITC) Amygdala neurone (25) has been adopted. Hypothalamus module in human brain receives information from input sensors and passes it to the Amygdala module to generate the fear.

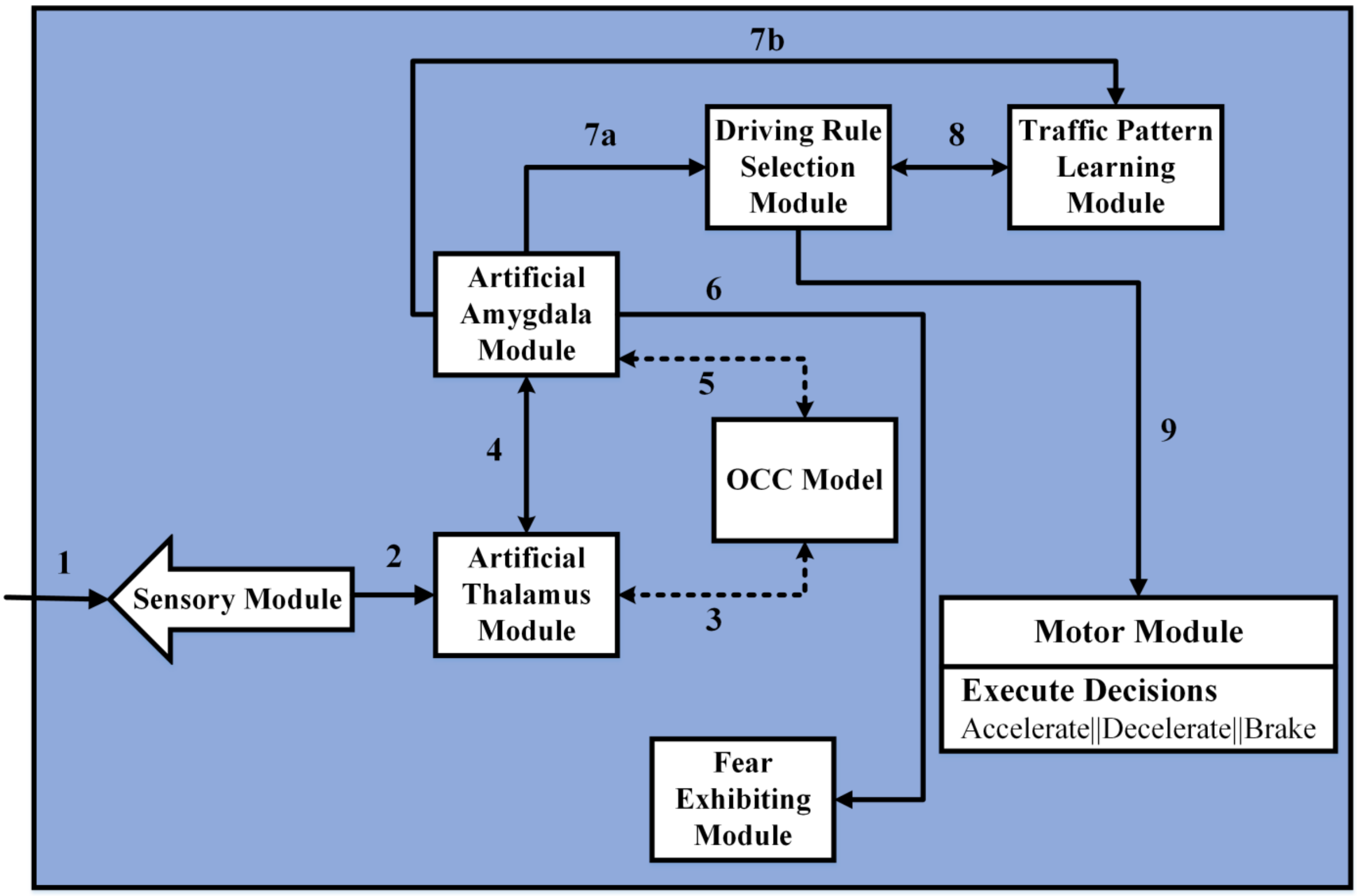

Figure. 2 EEEC_Agent based controller Architecture

It can be seen from the figure 2 that the proposed EEEC_Agent based controller architecture consists of seven main modules.

1 **Sensory Module**: It keeps track of the distance between neighbouring cars on a road segment.

2 **Artificial Thalamus Module**: The signals received by Thalamus module have different sonar frequencies that reflect on the prevailing inter-vehicular distance. On behalf of these sonar frequencies, Artificial Thalamus Module consults OCC model and compute the potential of fear emotion.

3 **OCC Model:** The OCC model act as an emotion generation engine by providing the emotions eliciting and emotions intensity computation mechanism.

4 **Artificial Amygdala**: This module computes the different intensity levels of fear, using prospect based emotion of OCC model, and pass different motion control instructions to the motor module accordingly.

5 **Driving Rule Selection Module:** This module helps the controller to select the driving rule according to the leading driving behaviour or the traffic pattern. We have proposed following three driving rules, which help the AV in avoiding the rear end collisions.

i. **IF** *Fear_Intenisty* is Low or Very Low

**THEN** Select Acceleration_Rate in High_Range AND Decleration_Rate in Low_Range

ii. **IF** *Fear_Intenisty* is Medium

**THEN** Select Acceleration_Rate in Low_Range AND Decleration_Rate in High_Range

iii. **IF** *Fear_Intenisty* is High or Very High

**THEN** Apply Brake

6 **Traffic Pattern Learning Module:** The module helps the AV to learn the behaviour of a leading human driver and then adapting the driving strategy accordingly. For example, if the following AV experience high number of brakes by the leading human driver then AV can learn that either the traffic is congested or the leading human driver is in aggressive behaviour. After learning this the AV will select the proper driving rule from the Driving Rule Selection Module. The rule given below help the EEEC_Agent to learn about the behaviour of leading driver or traffic pattern.

iv. **IF** No_of_Switches between *High and Medium fear* are frequent in a specified interval of time

**THEN** learn that leading driver is aggressive **OR** traffic pattern is busy so select rule no ii

**ELSE** Select rules from Driving Rule Selection Module according to the *Fear Intensity* level

7 **Motor Module**: This module acts in the place of the human driver to perform the accident avoidance manoeuvre by initiating AVs different actuators like brake and gas pedal.

### 3.2.1 Proposed Emotion Generation Mechanism and Functionality of EEEC_Agent

Referring to the figure 2, in step 1 the Sensory module keeps track of the distance between neighbouring AVs on a road segment. The module consists of different long and short range sonars, which provides distances from neighbouring vehicles. In step 2, the output of the sensory module, distance is passed to the Artificial Thalamus module, which acts similar to the hypothalamus module in the human brain. In step 3 these distances are then checked against the maximum allowed threshold to compute *fear potential* by computing likelihood, desirability and Ig variables (variables for computing prospect based emotions defined by OCC model).

In step 4, the fear potential is passed to the Artificial Amygdala module, which further computes the *fear intensity*, step *5*, with the help of OCC model. In step 6a, agent exhibits its different fear intensity levels, which help the agent to express its fear state to other neighbouring agents and help them to adapt their strategies accordingly. In step 7a, the controller selects the suitable driving rule according to the level of computed fear intensity. In step 7b, EEEC_Agent learn the behavior of leading human driver or traffic pattern according to the rule number 4 and select the suitable driving rule from Driving Rule Selection Module in step 8. In step 9, the selected driving rule is passed to the motor module, which in turn executes it using suitable actuator..

### 3.2.2 Proposed Emotions Intensity Computation Mechanism Using Fuzzy Logic

The term fuzzy logic was first coined by Lotfi A. Zadeh (26) in 1965 to solve the problem of modelling approximated real world mechanisms. The classic set theory deals with binary logic and hence fails to model real life fuzzy or approximated mechanisms. As human emotions are fuzzy and complex in nature, then using fuzzy sets for modelling the human emotions can be a suitable choice (27). Previously in literature, fuzzy logic has been utilised to model emotions in (28) (29). However, to our best knowledge fuzzy logic has not been utilised in the literature previously to compute the different intensity levels of prospect based emotions. So, to compute the different intensity levels of prospect based emotion, fear in our case, experiments have been conducted in the next subsection.

## 4. Experiments

We have performed three different types of experiments. The first type of experiments help in computing the different intensity levels of fear emotion. The second type of experiments helps in simulation based verification of fear generation mechanism of EEEC_Agent using agent based simulation tool i.e. Netlogo and the third type of experiments have been performed to practically validate the Netlogo simulation results, proposed rear end collision avoidance rules, and tweak handling using specially built prototype AV platform.

## 4.1 Experiments Type 1: Computing the Different Intensity Levels of Fear Emotion Using Fuzzy Logic

To compute the different intensity levels of fear, we have built a Mamdani fuzzy inference system, which uses the traceability algorithm defined in (21). The computation traceability algorithm is given as under.

***If Prospect (v, e, t) and Undesirable (v, e, t) < 0***

***Then set Fear-Potential (v, e, t) =*** $f_f$ ***[|Desire (v, e, t) |, Likelihood (v, e, t), Ig (v, e, t)]***

***If Fear-Potential (v, e, t) > Fear-Threshold (v, t)***

***Then set Fear-Intensity (v, e, t) = Fear-Potential (v, e, t) - Fear-Threshold (v, t)***

***Else set Fear-Intensity (v, e, t) =0***

In order to compute the Fear-Potential of the computation traceability algorithm, we have to calculate the values of Desirability, Likelihood, and the Intensity of global (Ig) variables. In next three sub experiments A, B, and C we have computed the Desirability, Likelihood, and the Intensity of global (Ig) variables.

### a) Experiment A: Computing Undesirability

According to the OCC model, undesirability is a local variable which affects only event and agent based emotions. The undesirability variable further comprises two sub variables: 1) Importance of Goal (ImpGoal) 2) Achievement of Goal (AchGoal).

The effects of these two sub-variables in computing the desirability can be seen in the following scenario. Suppose that an AV is following a human driver on a highway and the goal of AV is reaching its destination without any rear end collision. Suddenly the leading human driven vehicle starts decelerating and the distance between two vehicles starts decreasing. Now here the undesirability of the event can have multi-values. If the speed of AV is very low and distance between both vehicles is very high, then the importance of goal, which is safety will be low and achievement of goal will be high because it is safe due to low speed and high distance. However, in another case if the AV is traveling at high speed and the distance between both vehicles is low, then the importance of goal will automatically set to high and the value of achievement of goal will depend on a quick and optimized response. The main simulation screen of computing desirability (undesirability in the case of fear) is shown in Fig. 3. The screen is showing two input variables and one output variable. The input variables are the importance of Goal (Imp Goal), achievement of the goal (AchGoal) and the output variable is Undesirability.

To compute Undesirability Mamdani model with centroid defuzzification has been utilised. Furthermore, the trigonometric function (trimf) as shown in equation 1 has been used as a membership function.

$$f(\boldsymbol{\omega}, \boldsymbol{d}, \boldsymbol{e}, \boldsymbol{f}) = \begin{Bmatrix} \mathbf{0} & \boldsymbol{\omega} \leq \boldsymbol{d} \\ \mathbf{1} - \boldsymbol{d}/_{\boldsymbol{e} - \boldsymbol{d}} & \boldsymbol{d} < \omega \leq e \\ \boldsymbol{f} - \boldsymbol{\omega}/_{\boldsymbol{f} - \boldsymbol{e}} & \boldsymbol{e} < \omega \leq f \\ \mathbf{0} & \boldsymbol{\omega} > f \end{Bmatrix} \quad (1)$$

Where **d** and **f** represent the feet of the triangle, **e** is the peak of the triangle and **f** **($\boldsymbol{\omega}$)** represent the membership of the variable $\boldsymbol{\omega}$**.** Furthermore, domain of each variable is divided into 5 parts.

We have used fuzzy rules to infer the undesirability of collision event. For this purpose, ImpGoal and AchGoal have been considered as input variables as shown in the figure 3.

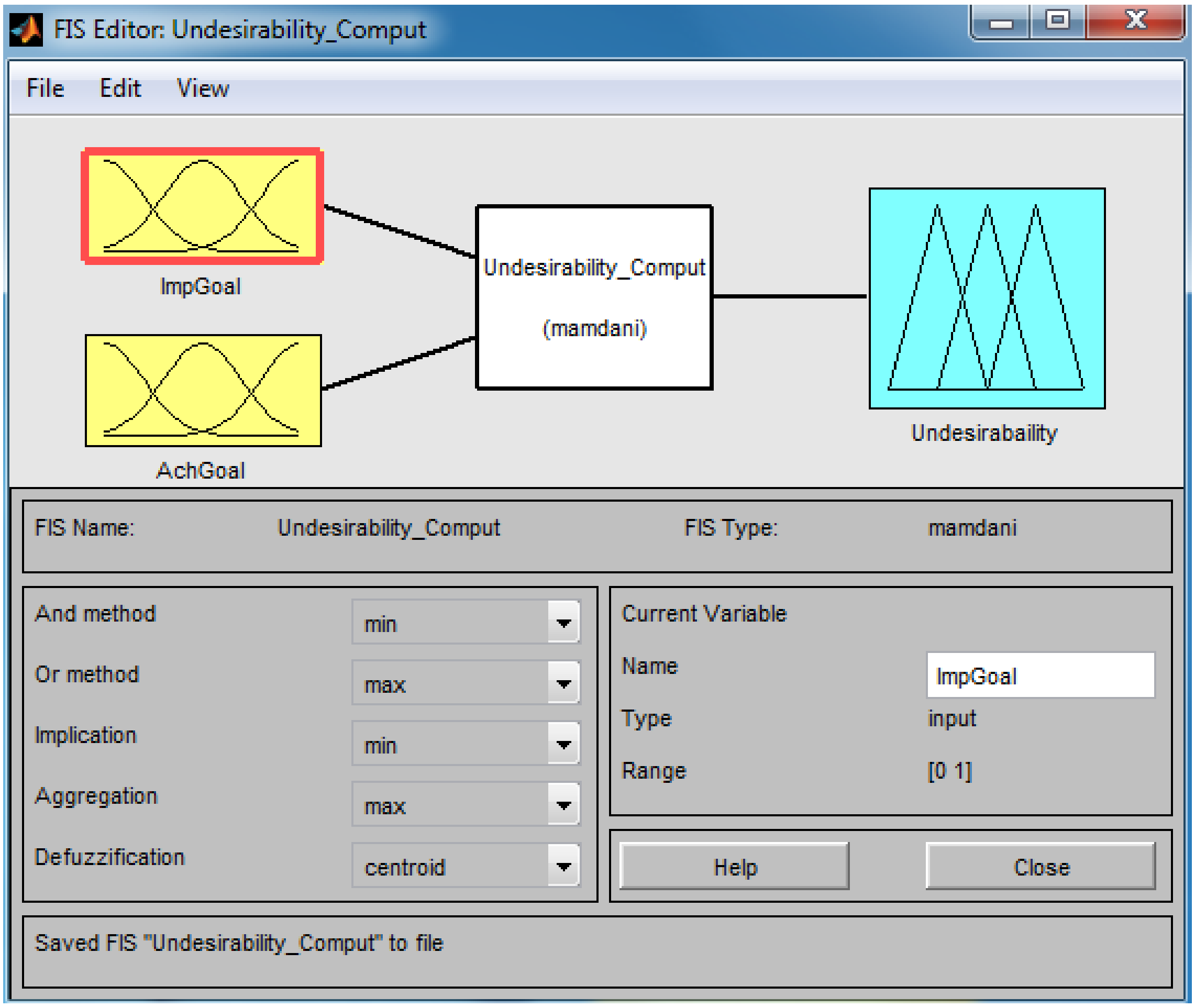

Figure. 3 Main simulation screen for Undesirability computation

The membership function of ImpGoal variable is described using five fuzzy sets: Very Low Importance of Goal (VLimpG), Low Importance of Goal (LImpG), Medium Importance of Goal (MImpG), High Importance of Goal (HImpG), and Very High Importance of Goal (VHImpG) as shown in figure 4a. The importance of a goal is dynamically set according to the agent's assessment of a particular situation. The membership function of AchGoal variable is represented by five fuzzy sets: No Achievement of Goal (NAG), Low Achievement of Goal (LAG), Medium Achievement of Goal (MAG), High Achievement of Goal (HAG), and Very High Achievement of Goal (VHAG) as shown in figure 4b. Finally, the membership function of Undesirability variable can be described as Very Low Undesirable (VLUD), Low Undesirable (LUD), Medium Undesirable (MUD), Highly Undesirable (HUD), and Very Highly Undesirable (VHUD) as shown in figure 4c.

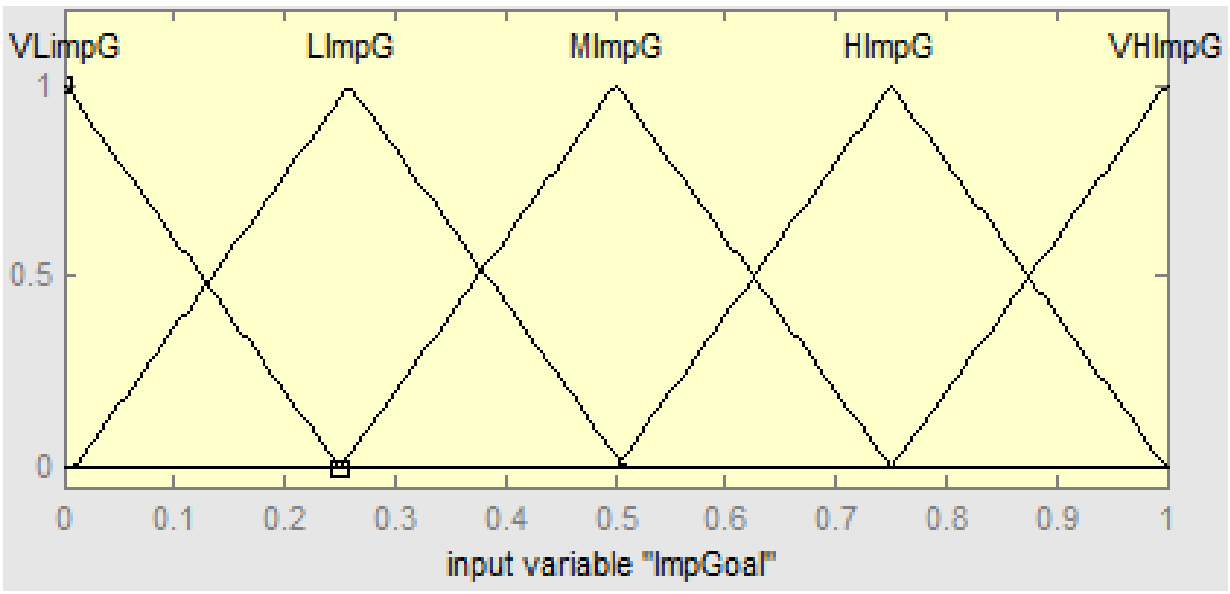

**(a)**

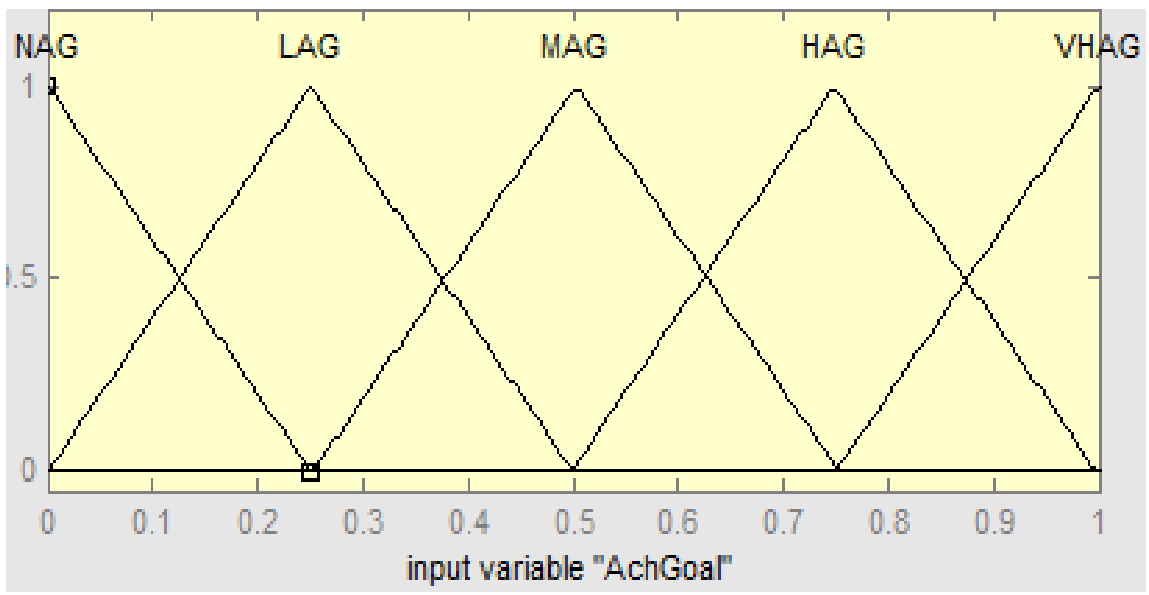

**(b)**

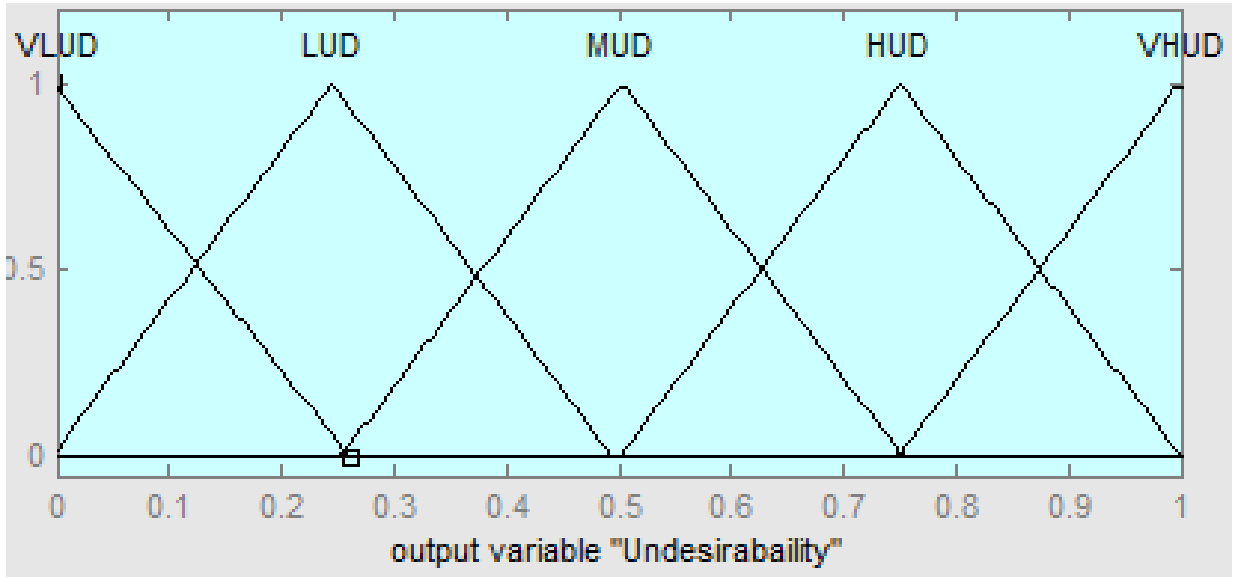

**(c)**

Figure. 4 Membership functions. (a) ImpGoal variable , (b) AchGoal variable, (c) Undesirability variable

To determine the Undesirability of events based on the importance of goal and achievement of goal, fuzzy rules of the form given below has been used:

IF Importance of (G) is A1

AND achievement of (G) is A2

THEN Undesirability of E is C

The detailed set of undesirability inference rules is presented in table 1.

Table 1. Detailed set of Undesirability Inference Rules

| If Importance of Goal is | And Achievement of goal is | Then undesirability will be |
|---|---|---|
| VLImpG | NAG | MUD |
| VLImpG | LAG | LUD |
| VLImpG | MAG | LUD |
| VLImpG | HAG | VLUD |
| VLImpG | HFAG | VLUD |
| LImpG | NAG | MUD |
| LImpG | LAG | MUD |
| LImpG | MAG | LUD |
| LImpG | HAG | VLUD |
| LImpG | VHFAG | VLUD |
| MImpG | NAG | HUD |
| MImpG | LAG | MUD |
| MImpG | MAG | MUD |
| MImpG | HAG | LUD |
| MImpG | VHFAG | LUD |
| HImpG | NAG | VHUD |
| HImpG | LAG | HUD |
| HImpG | MAG | HUD |
| HImpG | HAG | MUD |
| HImpG | VHFAG | VHUD |
| VHImpG | NAG | VHUD |
| VHImpG | LAG | HUD |
| VHImpG | MAG | HUD |
| VHImpG | HAG | HUD |
| VHImpG | VHFAG | MUD |

**Validation of fuzzy logic rules for computing the Undesirability:** The validation of undesirability fuzzy rules has been performed in rule viewer of FIS editor. Rule viewer was provided random values for different linguistic tokens and in result, fuzzy inference system computed different intensities of undesirability. To cross check, the outcomes hand tracing mechanism has been adopted, which further validated the outcomes of different undesirability values shown in Table 2. In test 1, it can be seen that input variables ImpGoal and AchGoal have values 0.1 and 0.5 which lie in the very low range and medium range respectively. In result, the FIS system computes low undesirability i.e. 0.25, which is correct. In the same way to test 7, the linguistic tokens medium importance of goal i.e. MImpG and medium achieved goal MAG have values 0.56 and 0.5 which lie in the medium range. In result, the FIS system computes medium intensity of undesirability i.e. 0.567, which is correct. In the same way, other validation results can be cross-checked using hand tracing mechanism.

Table 2 shows that different values for ImpGoal and AchGoal are entered as input and each time output value of the Undesirability variable is according to the rules.

Table 2. Validation of Undesirability Inference Rules

| No. Of Tests | ImpGoal | AchGoal | Undesirability |
|---|---|---|---|
| 1 | 0.1 (VLImpG) | 0.5 (MAG) | 0.25 (LUD) |
| 2 | 0.2 (VLImpG) | 1.0 (VHAG) | 0.08 (VLUD) |
| 3 | 0.27 (LImpG) | 0 (NAG) | 0.52 (MUD) |
| 4 | 0.30 (LImpG) | 0.5 (MAG) | 0.31 (LUD) |
| 5 | 0.4 (LImpG) | 1.0 (VHAG) | 0.09 (VLUD) |
| 6 | 0.5 (MImpG) | 0 (NAG) | 0.74 (HUD) |
| 7 | 0.56 (MImpG) | 0.5 (MAG) | 0.567 (MUD) |
| 8 | 0.6 (MImpG) | 1.0 (VHAG) | 0.09 (VLUD) |
| 9 | 0.8 (HImpG) | 0 (NAG) | 0.91 (VHUD) |
| 10 | 0.85 (HImpG) | 0.5 (MAG) | 0.746 (HUD) |
| 11 | 0.79 (HImpG) | 1.0 (VHAG) | 0.085 (VLUD) |
| 12 | 0.96 (VHImpG) | 0 (NAG) | 0.917 (VHUD) |
| 13 | 0.98 (VHImpG) | 0.5 (MAG) | 0.747 (HUD) |
| 14 | 1.0 (VHImpG) | 1.0 (VHAG) | 0.08 (VLUD) |

### b) Experiment B: Computing Likelihood

The Likelihood of the event depends on the Distance and Speed of the following and leading AVs. In our case Likelihood is representing TTA (Time to Avoid). For example, if the Distance between two vehicles is low and their Speed is in high range then it leads to higher Likelihood of collision between these two vehicles. So the two variables which affect the likelihood of an event are the Distance between both AVs and the Speed of Bullet AV. The main simulation screen for Likelihood computation is shown in figure 5.

The membership function of Distance variable is described using five fuzzy sets: Very Low Distance (VLD), Low Distance (LD), Medium Distance (MD), High Distance (HD), and Very High Distance (VHD) as shown in figure 6a. The distance between vehicles is dynamically changed according to the acceleration/ deceleration maneuvers of leading and following vehicles. In the same way, the membership function of Speed variable is represented by five fuzzy sets: Very Low Speed (VLS), Low Speed (LS), Medium Speed (MS), High Speed (HS), and Very High Speed (VHS) as shown in figure 6b. Finally, the membership function of Likelihood measure of event can be described as Very Low Likelihood (VLLH), Low Likelihood (LLH), Medium Likelihood (MLH), Highly Likelihood (HLH), and Very Highly Likelihood (VHLH) and has been presented in figure 6C.

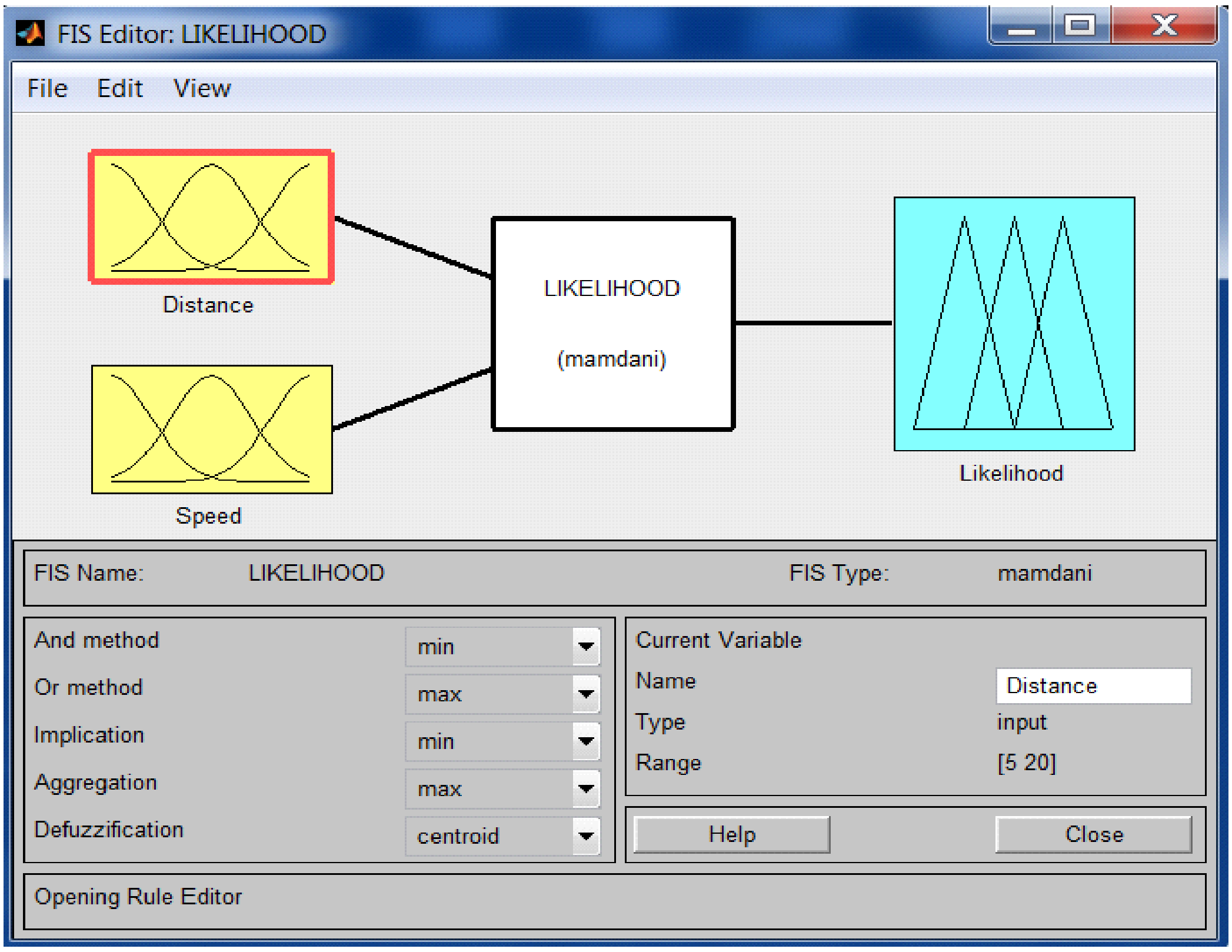

Figure. 5 Main simulation screen of Likelihood computation

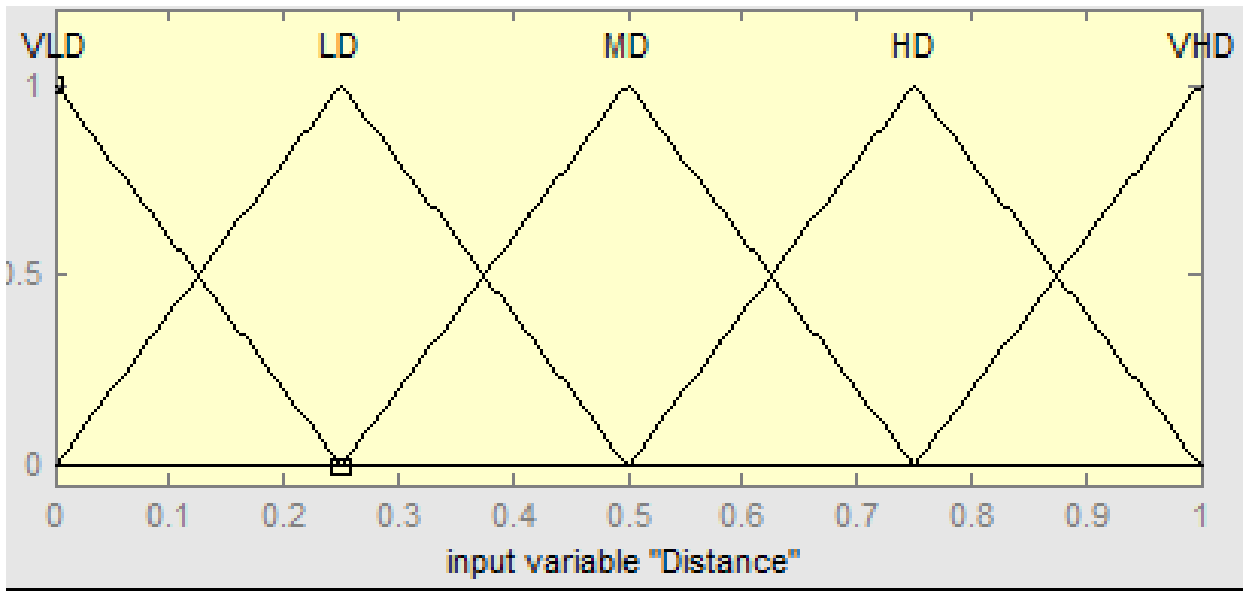

**(a)**

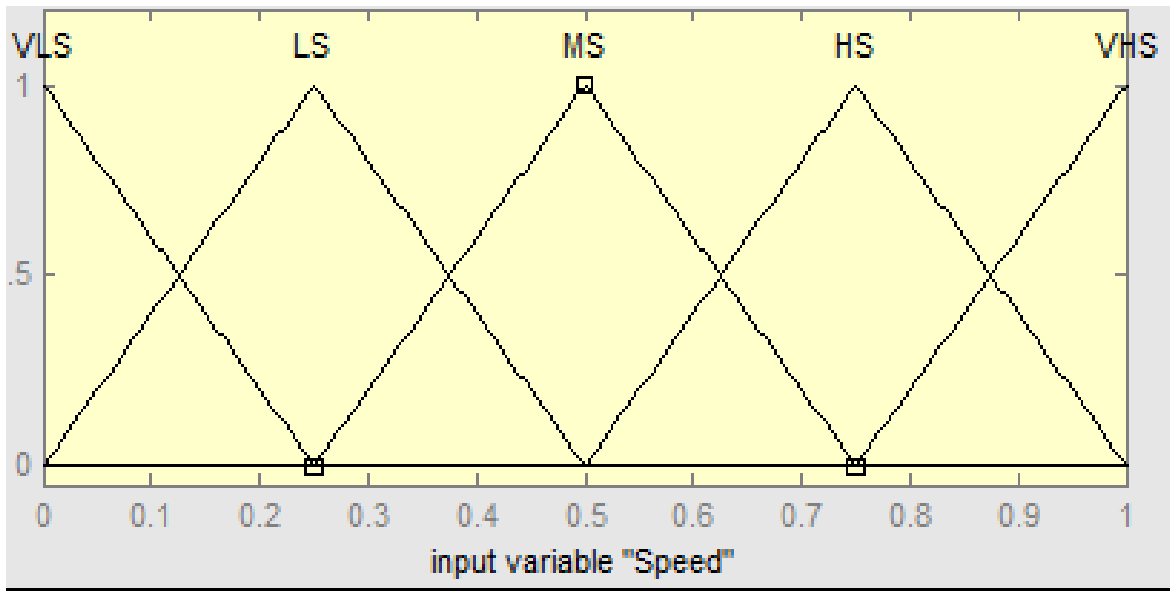

**(b)**

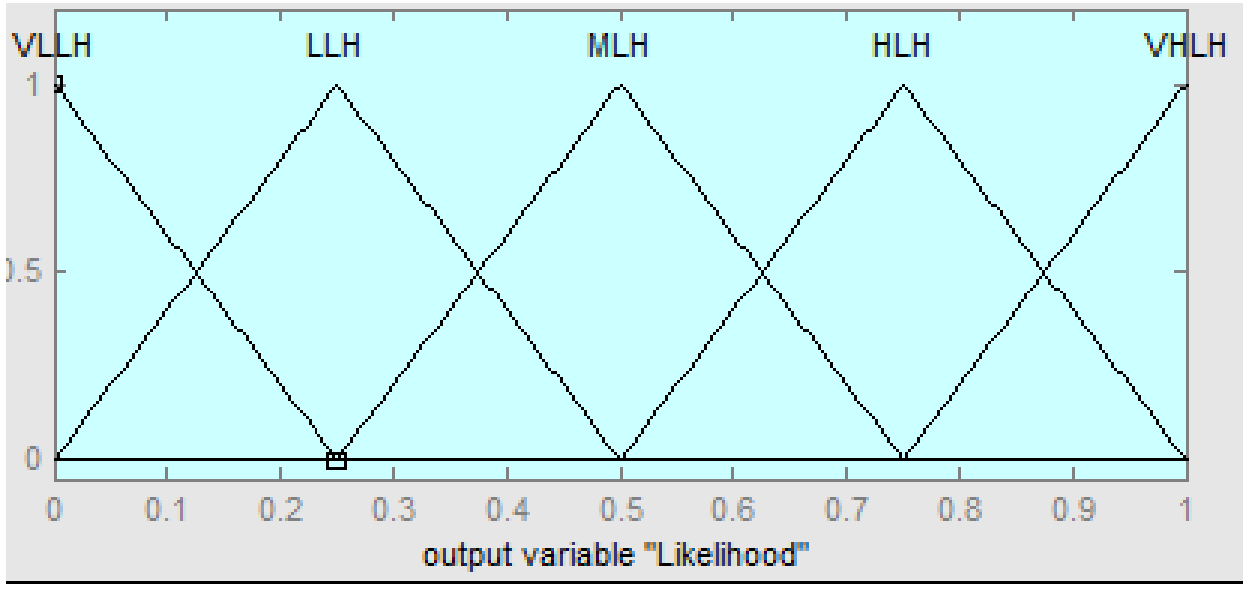

**(c)**

Figure. 6 Membership functions. (a) ImpGoal variable , (b) AchGoal variable, (c) Undesirability variable

To determine the likelihood of events based on distance and speed, we used fuzzy rules of the form given below:

IF Distance is A1

AND speed is A2

THEN Likelihood of E is C

Twenty-five rules were defined to obtain the value of the variable likelihood. These rules are given in the table 4.

Table 3. Likelihood Inference Rules

| If Distance is | And Speed is | Then Likelihood is |
|---|---|---|
| VHD | VHS | MLH |
| VHD | HS | LLH |
| VHD | MS | VLLH |
| VHD | LS | VLLH |
| VHD | VLS | VLLH |
| HD | VHS | HLH |
| HD | HS | MLH |
| HD | MS | VLLH |
| HD | LS | VLLH |
| HD | VLS | VLLH |
| MD | VHS | VHLH |
| MD | HS | VHLH |
| MD | MS | MLH |
| MD | LS | LLH |
| MD | VLS | VLLH |
| LD | VHS | VHLH |
| LD | HS | VHLH |
| LD | MS | HLH |
| LD | LS | MLH |
| LD | VLS | VLLH |
| V LD | VHS | VHLH |
| V LD | HS | VHLH |
| V LD | MS | VHLH |
| V LD | LS | HLH |
| V LD | VLS | MLH |

Figure 4 is representing the main simulation screen utilised to compute the Likelihood variable. The screen is showing two input variables and one output variable. The input variables are Speed and Distance and the output variable is Likelihood, as discussed above.

### c) Experiment C. Intensity of Global Variable

The intensity of global variable further depends on Proximity and the Sense of Reality (SoR) variables. The SoR variable actually depicts the scene interpretation performed by the sensing module of agent and it has the global influence on the intensity of emotions. Proximity is the distance between the AVs. The proximity influences the intensity of emotions which can involve future situations. We have taken proximity here in spatial terms.

For a variable of Ig, the five membership functions VLIG, LIG, MIG, HIG, and VHIG are defined which represent the Very low intensity of Low likelihood, Medium likelihood, High likelihood and Very High likelihood respectively, shown in Table 5.

Table 4. Ig Linguistic tokens

| Linguistic Tokens | Description |
|---|---|
| VLIG | Very low intensity of global variable |
| LIG | Low intensity of global variable |
| MIG | Medium intensity of global variable |
| HIG | High intensity of global variable |
| VHIG | Very high intensity of global variable |

The intensity of global variable depends on proximity and SoR variables. Twenty-five rules are defined to obtain the value of the variable likelihood. These rules are presented in Table 5.

Table 5. Ig Inference Rules

| If Sense of Reality is | And Proximity is | Then Intensity of Goal will be |
|---|---|---|
| VLSOR | About to | MIG |
| VLSOR | Going to | MIG |
| VLSOR | MChance | LIG |
| VLSOR | LChance | VLIG |
| VLSOR | NChance | VLIG |
| LSOR | About to | HIG |
| LSOR | Going to | MIG |
| LSOR | MChance | MIG |
| LSOR | LChance | LIG |
| LSOR | NChance | VLIG |
| MSOR | About to | HIG |
| MSOR | Going to | HIG |
| MSOR | MChance | MIG |
| MSOR | LChance | LIG |
| MSOR | NChance | VLIG |
| HSOR | About to | VHIG |
| HSOR | Going to | HIG |
| HSOR | MChance | MIG |
| HSOR | LChance | LIG |
| HSOR | NChance | VLIG |
| VHSOR | About to | VHIG |
| VHSOR | Going to | VHIG |
| VHSOR | MChance | HIG |
| VHSOR | LChance | HIG |
| VHSOR | NChance | MIG |

Figure 7 presents the main simulation screen regarding the quantitative computation of Ig variable. Here the SoR and proximity are acting as two input variables and Ig as the output variable.

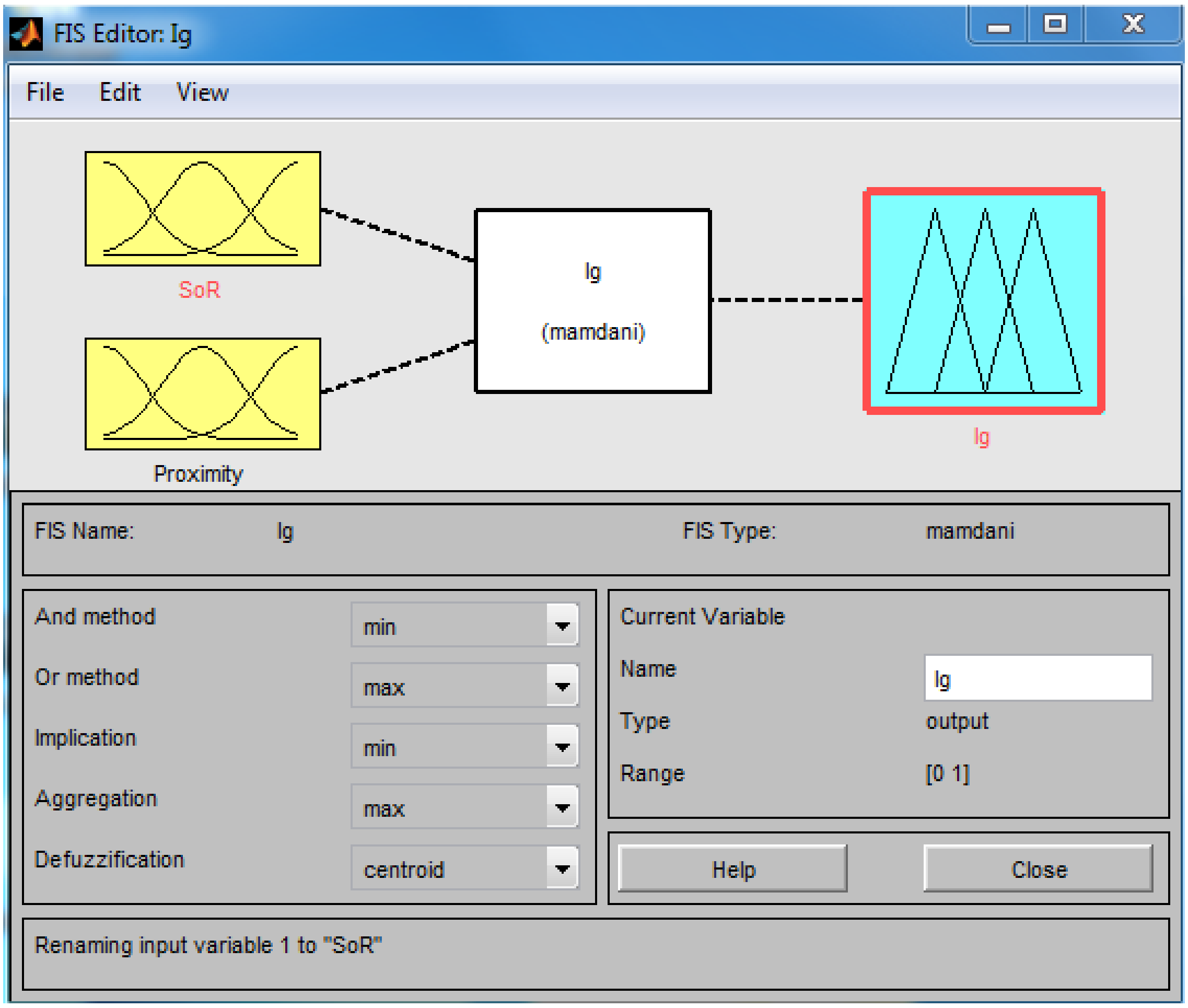

Figure. 7 Main Simulation Screen of the Intensity of global variable

## 4.2 Experiments Type 2: Simulation-Based Verification of Fear Generation Process of EEEC_Agent

The purpose of the second type of experiments is to verify the fear generation mechanism of proposed EEEC_Agent based controller in the car following model. The end results of these tests help us in tuning the performance of the EEEC_Agent based controller before testing it in a real car following test. For simulation study, NetLogo 5.3 has been utilised which is a standard agent-based simulation environment. The NetLogo 5.1 environment consists of patches and turtles.

### 4.2.1 Experimental Parameters

To perform the simulation experiments empirically, three types of simulation parameter sets are defined. The first set consists of numeric values (Table 8, 9 and 10) of prospect based emotions (i.e. Fear) variables like the likelihood of accident event, the undesirability of accident event, and Ig. The second set of parameters consists of Stopping Sight Distance and overtaking sight distance described by equations 2 and 3.

$$\mathrm{SSD} = 1.47\mathrm{Vt} + \frac{1.075\mathrm{V}^2}{\mathrm{a}} \quad (2)$$

Where SSD = Stopping Sight Distance in feet, V = design speed in mph, t = brake reaction time in seconds and a = deceleration rate, 11.2 ft/s2.

$$\mathrm{OSD} = \mathrm{V_b}\,\mathrm{t} + 2\mathrm{s} + \mathrm{V_b}\sqrt{\frac{4\mathrm{s}}{\mathrm{a}}} \quad (3)$$

Here $V_b$ = velocity of overtaking Vehicle, t = Reaction time, S=space before and after overtaking and a = Maximum overtaking acceleration at different speeds.

The third miscellaneous set of parameters is presented in table 6.

Table 6. Miscellaneous set of Simulation parameters and their description

| **Simulation General Parameters** | **Range** | **Description** |
|---|---|---|
| No_of_Bullet_AV | [1] | This slider sets the number of the following vehicles in a car following test. |
| No_of_Target_AV | [1] | This slider sets the number of the leading vehicles in a car following test. |
| Min_velocity_bullet | [0 -1; with increment of 0.1] | This slider helps in defining the lower boundary of velocity achieved by bullet vehicle in a car following test. |
| Max_velocity_bullet | [0 -1; with increment of 0.1] | This slider helps in defining the maximum velocity that can be achieved by bullet vehicle in a car following test. |
| Acceleration-bullet | [0 - 0.1; with increment of 0.01] | This slider helps in defining the maximum acceleration rate of bullet vehicle |
| Deceleration-bullet | [0 - 0.1; with increment of 0.01] | This slider helps in defining the minimum declaration- rate of bullet vehicle |
| Acceleration-target | [0 - 0.1; with increment of 0.01] | This slider helps in defining the maximum acceleration rate of the target vehicle |
| Deceleration-target | [0 - 0.1; with increment of 0.01] | This slider helps in defining the minimum declaration- rate of the target vehicle |
| Separation | [1 -20-; with increment of 2] | This slider helps in setting different initial distances between the bullet and target vehicles. |

Figure 8 presents the experimental environment along with input and output parameters. Two AVs are taking part in this simulation. The bullet AV is acting as the following agent, whereas the second one is leading AV which acts as a target agent. The left side of the simulation world contains input sliders for providing fuzzy logic based numeric values of prospect based emotion variables (Undesirability, Likelihood, and Ig). It is important to recall here that these numeric values of prospect based emotions were computed through experiments A), B) and C), presented in section 6.1, using fuzzy logic and then provided to the agent based simulation. The world size used in the simulation is (-25, -25) to (25, 25) and in this way, the total number of patches in the world is 25. To map the real-world distance in feet on 25 patches, each patch is representing a value equal to 100 feet.

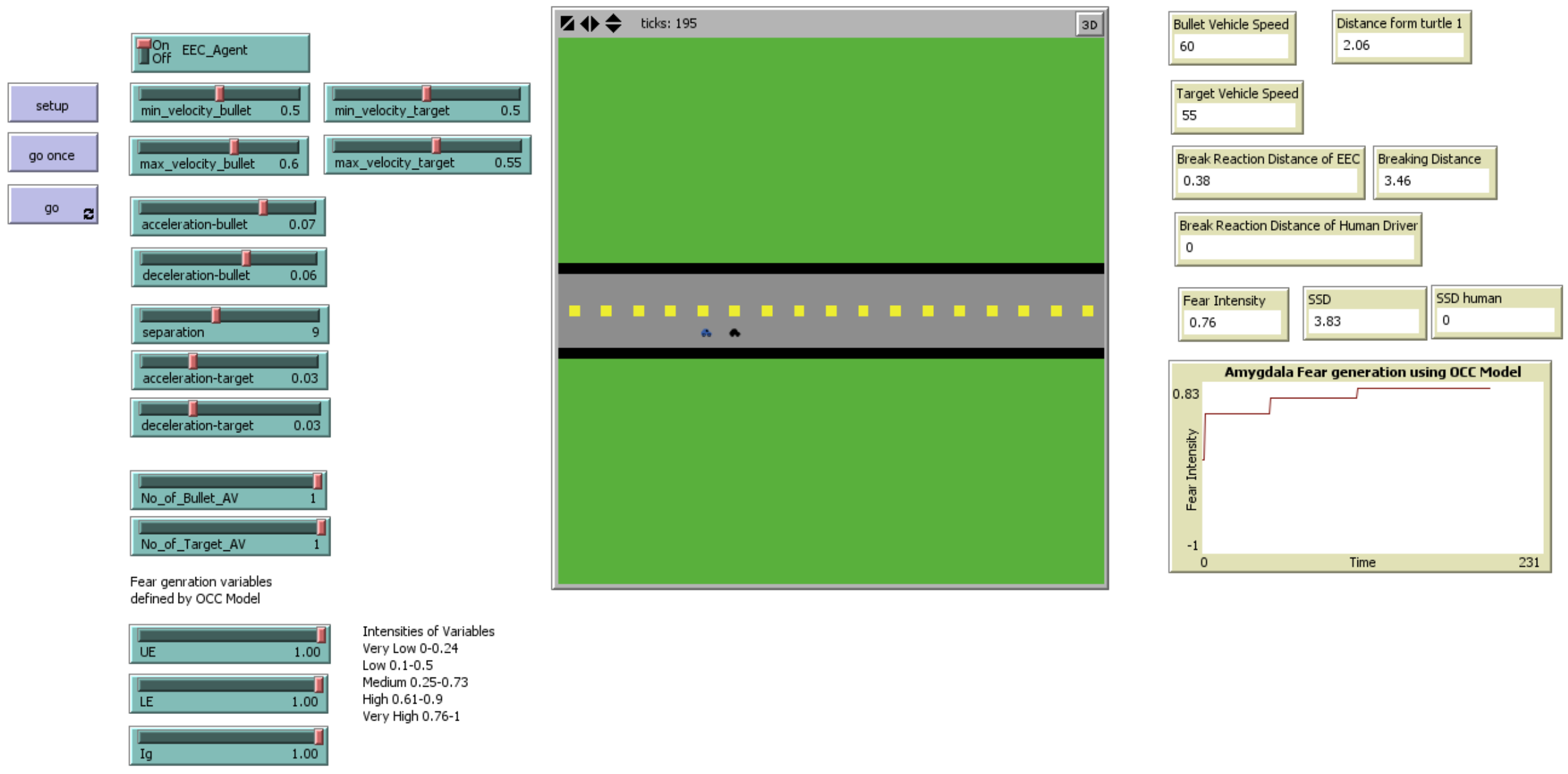

Figure. 8 Main Simulation Screen of EEEC_Agent based Collision Avoidance system in NetLogo Environment

### 4.2.2 Car Following Experiments Design

The basic purpose of the simulation is to verify the proposed fear generation mechanism of EEEC_Agent in a car following mode. For this purpose, six different types of the car following tests with different values have been designed as presented in Table 7. These tests are set up to check that if the distance between both AVs will be decreased, then what will be its effect on the intensity of the fear of EEEC_Agent. This set of tests has been performed using the behaviour space tool of NetLogo 5.3.1 environment and each test has been repeated 50 times.

Table 7. Car following Tests

| **Experiment No** | **Separation** | **Min Velocity** | **Max Velocity** | **Acceleration_Bullet** | **Deceleration_Bullet** | **Acceleration_Target** | **Deceleration_Target** | **EEC_Agent** |
|---|---|---|---|---|---|---|---|---|
| 1 | 5 | 10 | 100 | 0.06 | 0.03 | 0.03 | 0.03 | True |
| 2 | 9 | 10 | 100 | 0.06 | 0.06 | 0.03 | 0.03 | True |
| 3 | 13 | 10 | 100 | 0.06 | 0.03 | 0.03 | 0.03 | True |
| 4 | 13 | 60 | 100 | 0.06 | 0.06 | 0.03 | 0.03 | True |
| 5 | 17 | 10 | 100 | 0.06 | 0.06 | 0.03 | 0.03 | True |

# 5 Results and Discussion

This section describes the results of both experiments 1 and 2. The results are compared with the state of the art EEEC_Agent proposed in (20).

## 5.1 Experiments Type 1

Table 8 shows the quantitative values of undesirability from very low (VL) to very high (VH). The terms VLD, LD, MD, HD, and VHD represent very low desirability, low desirability, medium desirability, high desirability and very high desirability respectively. If the agent has a value between 0-0.24 for its undesirability of an event, then it can be interpreted as the very low undesirability. However, from an abstract analysis, it can be noted that due to the fuzzy nature of the emotion fear, the boundary of one intensity level mixes in the boundary of another intensity level. Hence, the intensity levels lying between 0.24 and 0.5 will be interpreted as low undesirability and lower than these values as the very low undesirability. In the same way, the other intensity levels of undesirability variable can be interpreted.

In the same way, Table 9 and Table 10 are presenting the five quantitative values for finding the different intensity levels of likelihood and Ig variables.

These quantitative values of Desirability, Likelihood and Ig are presented in Table 8, 9 and 10 respectively. These values are then provided to the EEEC_Agent for computing different intensities of fear in the next section by following the proposed SimConnector design.

Table 8. Quantitative Values of Five Intensity levels of Desirable Variable

| VLD | LD | MD | HD | VHD |
|---|---|---|---|---|
| 0-0.24 | 0.1-0.5 | 0.25-0.73 | 0.51-0.9 | 0.76-1 |

Table 9. Quantitative Values of Five Intensity levels of Likelihood Variable

| VLL | LL | ML | HL | VHL |
|---|---|---|---|---|
| 0-0.24 | 0.1-0.5 | 0.25-0.73 | 0.51-0.9 | 0.76-1 |

Table 10. Quantitative Values of Five Intensity levels of Global Variable (Ig)

| VLIg | LIg | MIg | HIg | VIg |
|---|---|---|---|---|
| 0-0.24 | 0.1-0.5 | 0.25-0.73 | 0.51-0.9 | 0.76-1 |

## 5.2 Experiments Type 2

Verification of fear generation mechanism of proposed EEC agent has been provided through an extensive set of tests over different arrangements of experiments. As five different sets of experiments have been designed considering short to long distances, separations, between Bullet and Target AVs. In Table 11 results have been given for validating the EEC agent by placing Bullet and Target AVs 5 separation apart. Bullet Vehicle is moving at a low speed of 10 mph and accelerating at a rate of 0.06 mph and decelerating with a low rate of 0.03 mph. At tick number 1, which is shown by the first record in Table 11, Bullet vehicle requires 0.16 feet SSD. Whereas the actual distance, i.e. 6.28 feet between vehicles is greater than the required SSD. That is why medium level fear is felt by EEC agent i.e. 49. As the Bullet vehicle proceeds further by adding 0.06 mph to its current speed, a decrease in the distance has been recorded which is shown by the second entry of distance in Table 11 i.e. 5.77 feet. This decrease in distance increased fear intensity and shifted it to the high level. As Bullet vehicle continues to accelerate the require SSD varies with changes in speed on every tick. The fourth record in Table 11 is showing the status of Bullet AV with an increased SSD value which is 4.73 due to increasing its speed. At this point, the autos' separation has crossed the safety sight distance limit, which is causing our EEC agent to feel high positive fear i.e. 76. After the violation of SSD, Bullet vehicle tends to decelerate. The rest of the entries are confirming the fact that deceleration causing an increase in distance and an ultimately decrement in fear level. Graphical representation of all 100 ticks' data has been provided in figure 9.

Table 11. Computation of Fear at low speed of Bullet AV with separation 5

| Speed=10 | Acceleration= 0.06 | Deceleration= 0.03 |
|---|---|---|
| **SSD** | **Distance** | **Fear Intensity** |
| 0.16 | 6.28 | 49 |
| 2.61 | 5.77 | 66 |
| 3.95 | 4.60 | 66 |
| 4.73 | 3.83 | 76 |
| 10.22 | 7.61 | 76 |
| 10.22 | 11.98 | 66 |
| 10.22 | 14.15 | 49 |
| 10.22 | 18.15 | 36 |
| 10.22 | 17.50 | 36 |

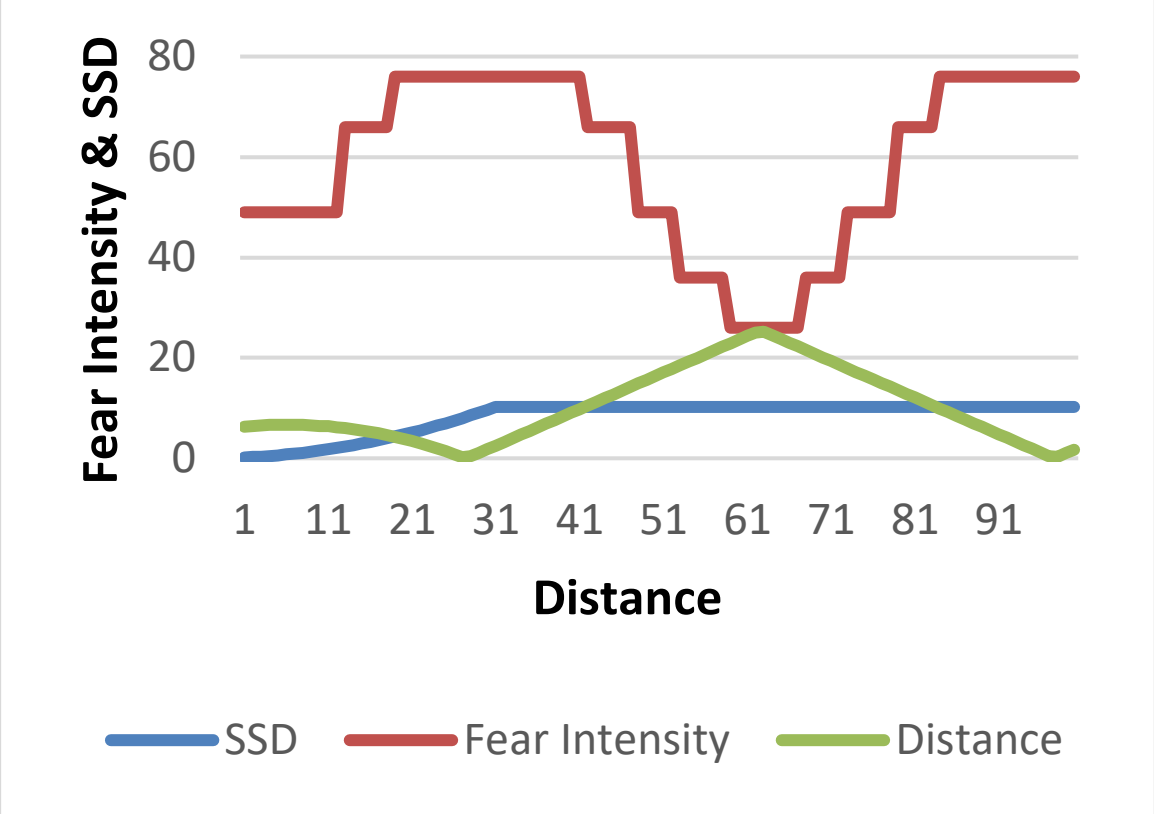

Figure. 9 Computation of Fear at low speed of Bullet AV With high acceleration and low deceleration

Table 12. Computation of Fear at low speed of Bullet AV with separation 9

| Speed=10 | Acceleration= 0.06 | Deceleration= 0.06 |
|---|---|---|
| **SSD** | **Distance** | **Fear Intensity** |
| 0.16 | 10.24 | 36 |
| 0.16 | 12.21 | 26 |
| 0.16 | 13.27 | 26 |
| 0.16 | 16.44 | 16 |
| 0.16 | 19.49 | 16 |
| 0.16 | 19.60 | 16 |
| 0.16 | 20.14 | 160.15 |
| 0.16 | 21.87 | 6 |

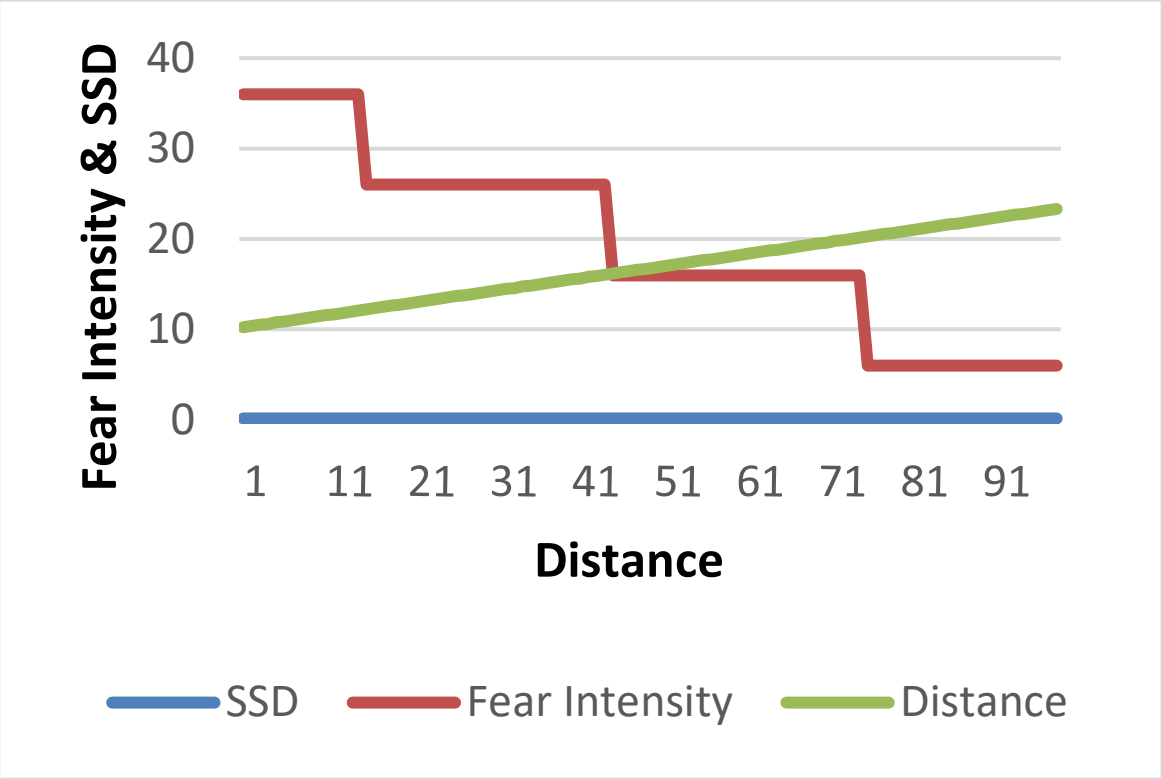

Figure. 10 Computation of Fear at low speed of Bullet AV With equally high acceleration and deceleration

Table 13. Computation of Fear at high speed of Bullet AV with separation 17

| Speed=10 | Acceleration= 0.06 | Deceleration= 0.06 |
|---|---|---|
| **SSD** | **Distance** | **Fear Intensity** |
| 0.15 | 18.37 | 16 |
| 0.15 | 20.21 | 6 |
| 0.15 | 22.44 | 6 |
| 0.15 | 24.12 | 6 |
| 0.15 | 25.35 | 6 |
| 0.15 | 20.32 | 6 |
| 0.15 | 19.88 | 16 |
| 0.15 | 18.92 | 16 |

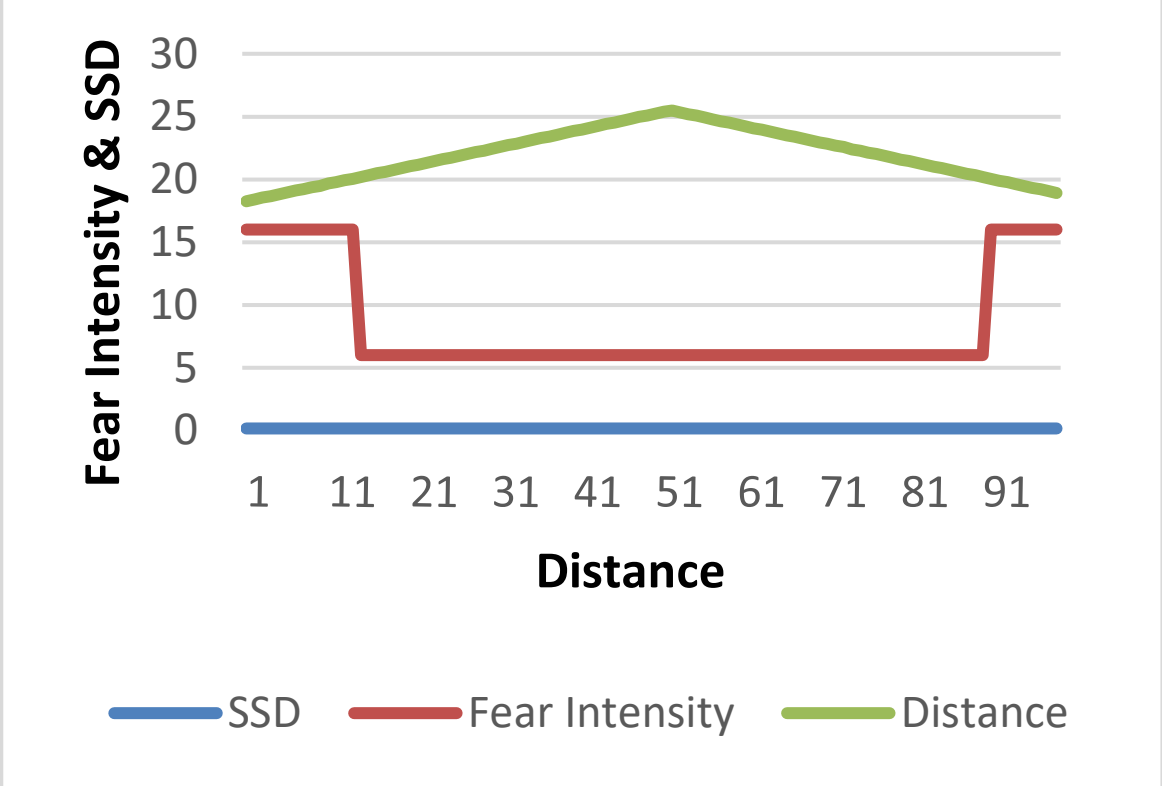

Figure. 11 Computation of Fear at high speed of Bullet AV with equally high acceleration and deceleration

Table 12 presents the figures regarding the validation test being performed with a low initial speed of Bullet vehicle, i.e. 10 mph with the initial separation of 9. Bullet vehicle has equally high acceleration and deceleration rate of 0.06 mph. Initial separation has been increased by 4 points than the previous setup. Staircase representation shown in figure 10 is substantiating the reality of the increase in the distance causes a decrease in the fear intensity. Required SSD for Bullet vehicle is 0.16 feet due to its low speed. Initial placement of the Bullet and Target vehicles is far apart. The distance shown by the first record in Table 12 is 10.24 feet, which has allowed Bullet AV to move continuously with mentioned rate by feeling positive low fear given by the value 36. An SSD column of Table 12 is baring a constant value due to an equal change in speed caused by both acceleration and deceleration. Deceleration of Bullet AV causing an increase in distance, for instance, entry number 3 shows the distance value of 13.27 feet hence lowering the fear value to 26. Fear continues to drop and that is because of the fact that Bullet AV is decelerating without any considerable fear intensity. Figure 10 shows a plot of 100 records, to show the complete behaviour of EEC agent shown through the course of this test.

The last subject of discussion regarding experiment tests is describing the results by putting vehicles distant apart, which is shown by the separation of 17 in the simulation. Bullet vehicle is travelling at a speed of 10 mph with equally high acceleration and deceleration rates i.e. 0.06 mph. The initial low speed of Bullet is causing an increase in distance of 18.37 feet to 20.21 feet, from 20.21 feet to 25.35 feet and hence lowering the fear from 16 to 6 (shown by record number 1 to 5 in Table 13). A gradual decrease in the distance and then a corresponding increase in the fear intensity have been portrayed through the rest of the records of Table 13. Figure 11 presents the actual behaviour of all 100 records that were gathered during 100 runs of the simulation. The downward movement of the orange line (fear) with the upward movement of the grey line (distance) is validating our claim regarding fear generation mechanism of EEEC agent.

# 6 Experiments Type 3. Practical validation of EEEC_Agent Based Collision Avoidance Controller

To verify our proposed controller, various types of practical validations have been performed. The first purpose of practical validations was to verify the simulation results regarding the fear generation process of proposed EEEC_Agent, the 2nd purpose was to validate the performance of EEEC_Agent based controller in terms of collision avoidance from human-driven vehicles, and third purpose was to check its performance during sudden or highly unexpected events like aggressive brake by leading human driven vehicle and sudden appearance of pedestrian or road hazard, which can be called as tweak handling.

The next subsection describes the step by step details of prototype AV test bed, which has been utilised to validate the EEEC_Agent controller in different cases.

## 6.1 Mapping EEEC Agent Based Controller Architecture To The Prototype AV Platform

The proposed EEEC_Agent based controller architecture, presented in the figure 2 has been mapped to the hardware architecture, figure 12, to build the AV test bed for rigorous validation of the EEEC_Agent based controller. The sensory module of EEEC_Agent has been mapped to the front, left, and right, high range ultrasonic sonars, which help in computing the distances from neighbouring human-driven vehicles. The thalamus module has been replaced with an Arduino Mega Board, which is based on the ATmega2560 processor. The input of ATmega2560 is provided to the C # program installed on Microsoft windows-10 based tablet, emulating artificial Amygdala unit, to compute the final intensity of emotions with the help of OCC model. Then, according to the intensity of fear, instructions regarding acceleration, deceleration and brakes are provided to the different actuators through another ATmega2560 processor, which act as a motor module of EEEC_Agent. It is important to mention that DC controller has been utilised between the motor module and actuators, which help in avoiding collisions with smooth acceleration and deceleration rate, within the passenger comfort level, along the abrupt brakes for tweaks handling. Figures 13a and 13b are presenting the prototype AV platform and practical AV-Human driven car following test respectively.

## A. Practical Validation Test To Validate The Simulation Results Regarding Fear Generation

In this practical validation, the simulation results regarding the fear generation process of EEEC_Agent has been validated. The test configuration is given as follows.

(1) The initial distance between prototype AV and stationary road hazard is random (almost between 5 meters to 17 meters).

(2) The results of each test have been traced into a log file every millisecond and each test has been repeated 5 times.

**Results Discussion**- Figure 14 presents the results of practical validation tests regarding fear generation mechanism of the proposed EEEC agent. Figure 14 consists of two graphs, the first one is Fear Intensity Vs Time graph, which shows the different levels of fear experienced by prototype AV and the second graph is Distance Vs Time graph, which presents the change in distance between prototype AV and stationary road hazard.

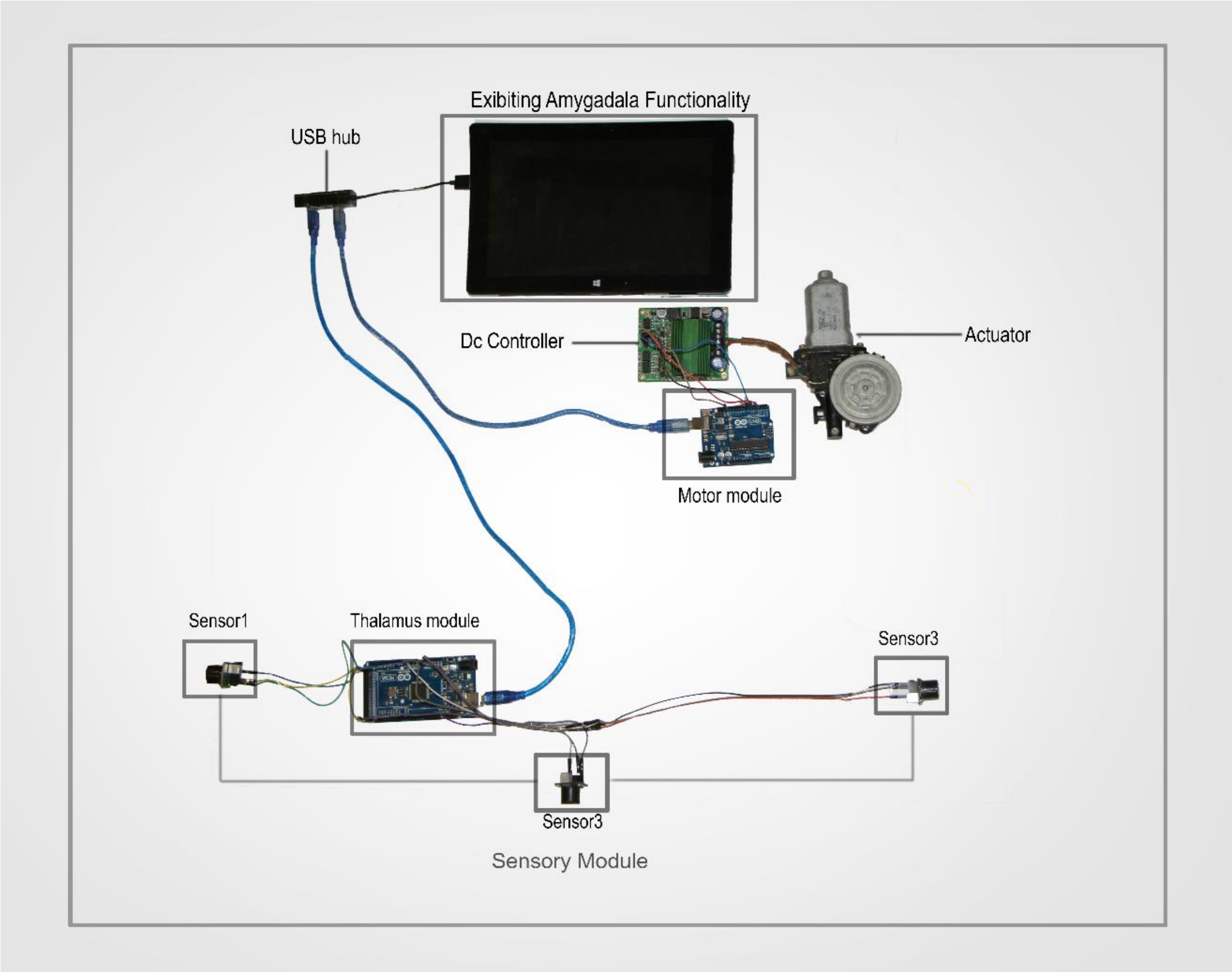

Figure. 12 Computation of Fear at high speed of Bullet AV with equally high acceleration and deceleration

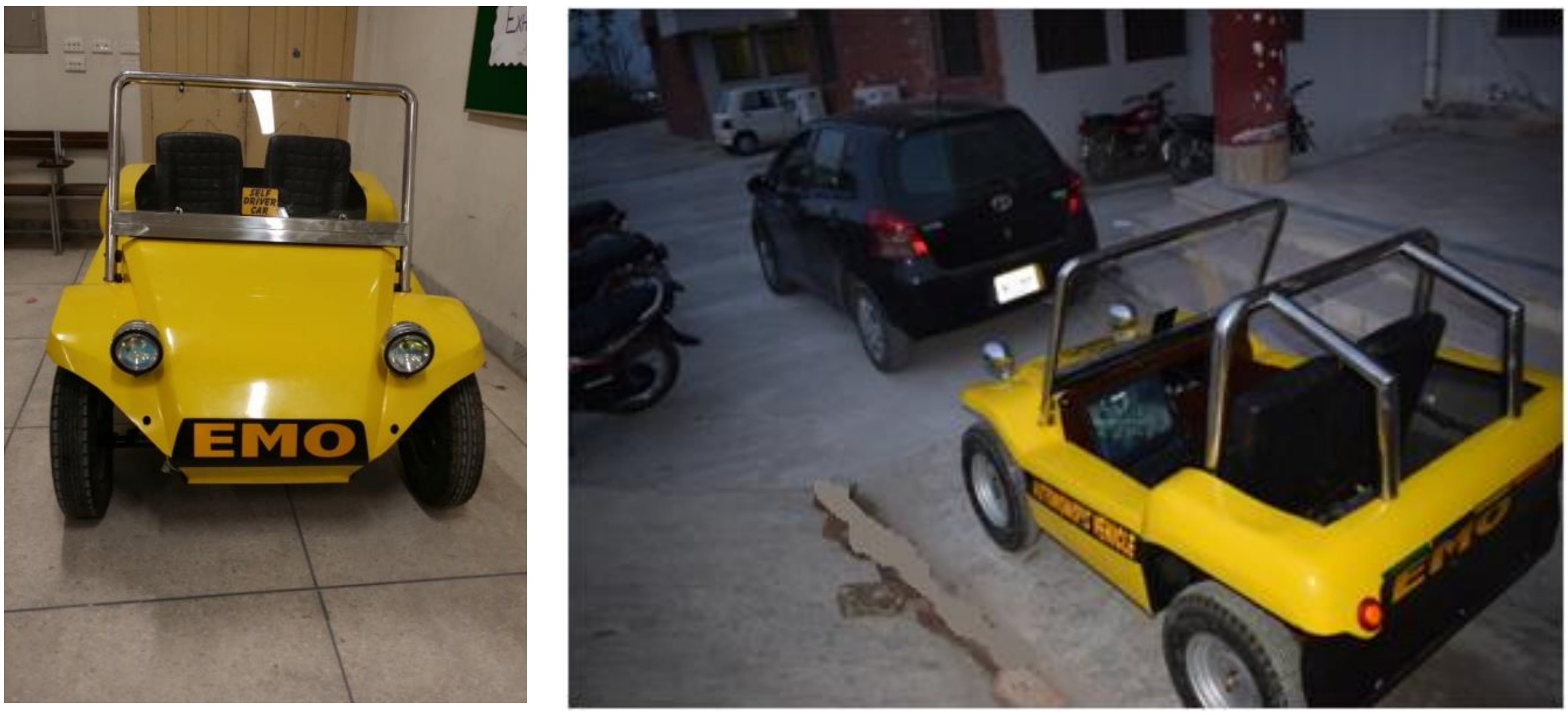

(a) (b)

Figure. 13 (a) EEEC_Agent Based Controller Installed prototype AV, (b) AV-Human driven Car following test

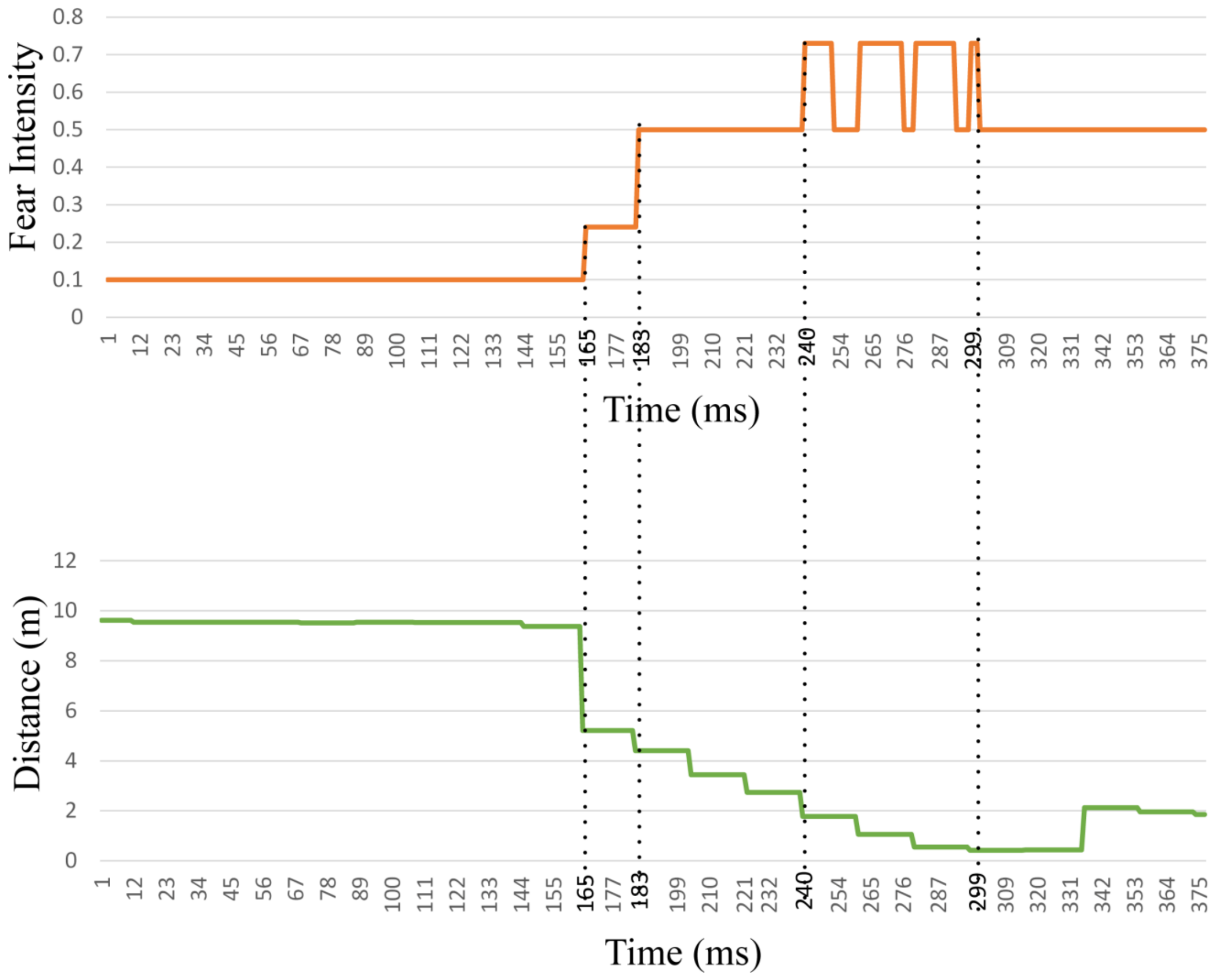

Figure. 14 Practical validation results of EEC_Agent's Fear Generation Mechanism

If we see the Fear Intensity graph, then it can be seen that up to 164 ms the prototype AV experienced very low fear i.e. 0.1 because the Distance graph depicts that for 164 ms the distance between prototype AV and stationary road hazard is almost 10 meters. However, at 165 ms the distance between prototype AV and stationary road hazard became less than 6 meters and the fear intensity level of prototype jumped from 0.1 to 0.25 i.e. Low level of Fear. In the continuation of the experiment, at 183 ms the distance between prototype AV and road hazard decreased up to 4 meters and in reaction, the fear intensity level of prototype jumped to the 0.5 i.e. medium level of fear. If we further see the timeline, then at 240 ms the distance reached to the 2 meters, high chances of a collision, which increased the fear intensity level of the prototype from 0.5 to 0.71 i.e. very high level of fear. Between 240 ms to 299 ms, the AV experienced very high fear due to the low distance between itself and potential collision threat. From these results, our simulation results regarding fear generation mechanism, presented in figures 9-11, of EEEC_Agent has been validated.

## B. Performance Measurement of EEEC_Agent Based Controller For The Rear End Collision Avoidance Using AV-Human Driven Car Following Test

In this practical validation, the rear end collision avoidance capabilities of EEEC_Agent using proposed driving rules have been validated. The test configuration is given as follows.

(1) The initial distance between two trailing and leading vehicles is 7 meters. The initial speed of human-driven leading car and trailing prototype AV is random between 1 meter/s to 3 meters/s.

(2) The human driven vehicle changes its speed with random acceleration and deceleration pattern. Here, we have divided this acceleration/deceleration variation according to non-aggressive and aggressive drivers and during tests, it was the driver choice that he adapts non-aggressive behaviour or an aggressive one. The aggressive and non-aggressive behaviour has been distinguished on the basis of the high and low number of sudden brakes respectively.

(3) The results of each test have been traced into a log file every millisecond and each test has been repeated 5 times.

**Results Discussion**- Figure 15 presents the behaviour of EEC_Agent in a car following test, when the leading human driver is in an aggressive mode or the traffic pattern is busy. From the time line of Fear Intensity Vs Time graph, it can be seen that for first 141 ms the prototype AV experienced low fear and it selects driving rule 1 with high acceleration, i.e. 0.05 $m/s^2$ and low deceleration pattern i.e. 0.03 $m/s^2$ as presented in Acceleration Vs Time and Deceleration Vs. Time graphs respectively. At 204 ms AV experienced very high fear and applied the brake by selecting rule no 3, as depicted, 0 speed at 205 ms, in the Speed Vs. Time graph. In the continuation of the analysis, if we see the time line of fear intensity graph, then it can be seen that between 477 ms to 1490 ms the AV experienced the frequent switches between high fear and medium fear. From these high numbers of switches EEC_Agent learned that leading human driver is driving aggressively or traffic pattern is busy then it automatically adopted low acceleration rate i.e. 0.03 $m/s^2$ and high deceleration rate i.e. 0.05 $m/s^2$. From the deceleration graph it can be seen an interesting capability of proposed EEC_Agent that it keeps the high deceleration and low acceleration pattern for next few milliseconds because of frequent switching between medium and high fear intensity levels and it learned that it is suitable to be in safe mode because of more chances of aggressive brakes by the leading human driver. If we further study the timeline between 1691ms to 2040 ms then it can be seen that AV experienced medium fear for 51 ms and in result, it picked the driving rule

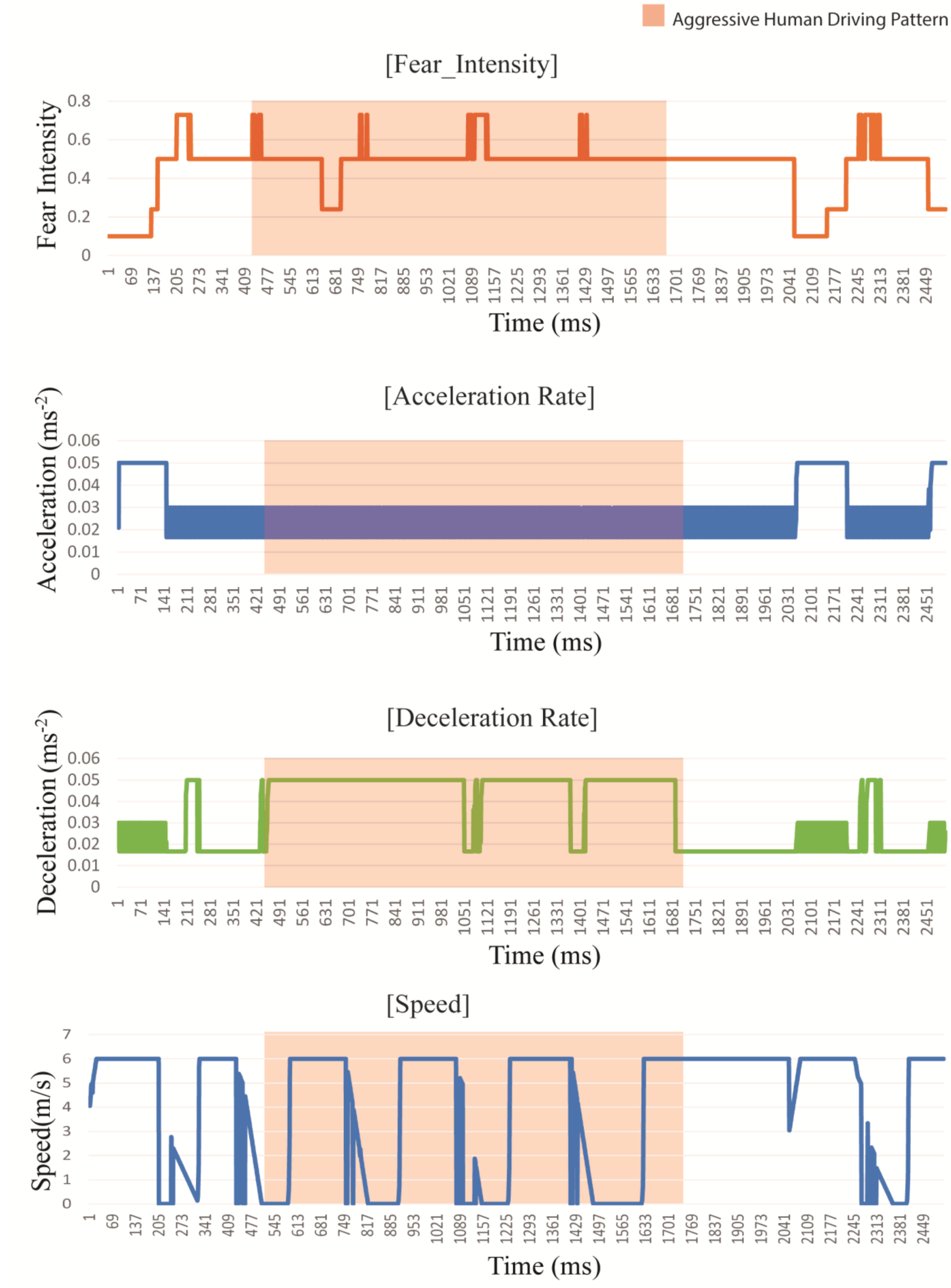

Figure. 15 Graphical depiction of results regarding performance measurement of EEEC_Agent based controller for rear end collision avoidance using AV-Human driven Car following test, when the leading human driver is in more aggressive mode or traffic pattern is busy

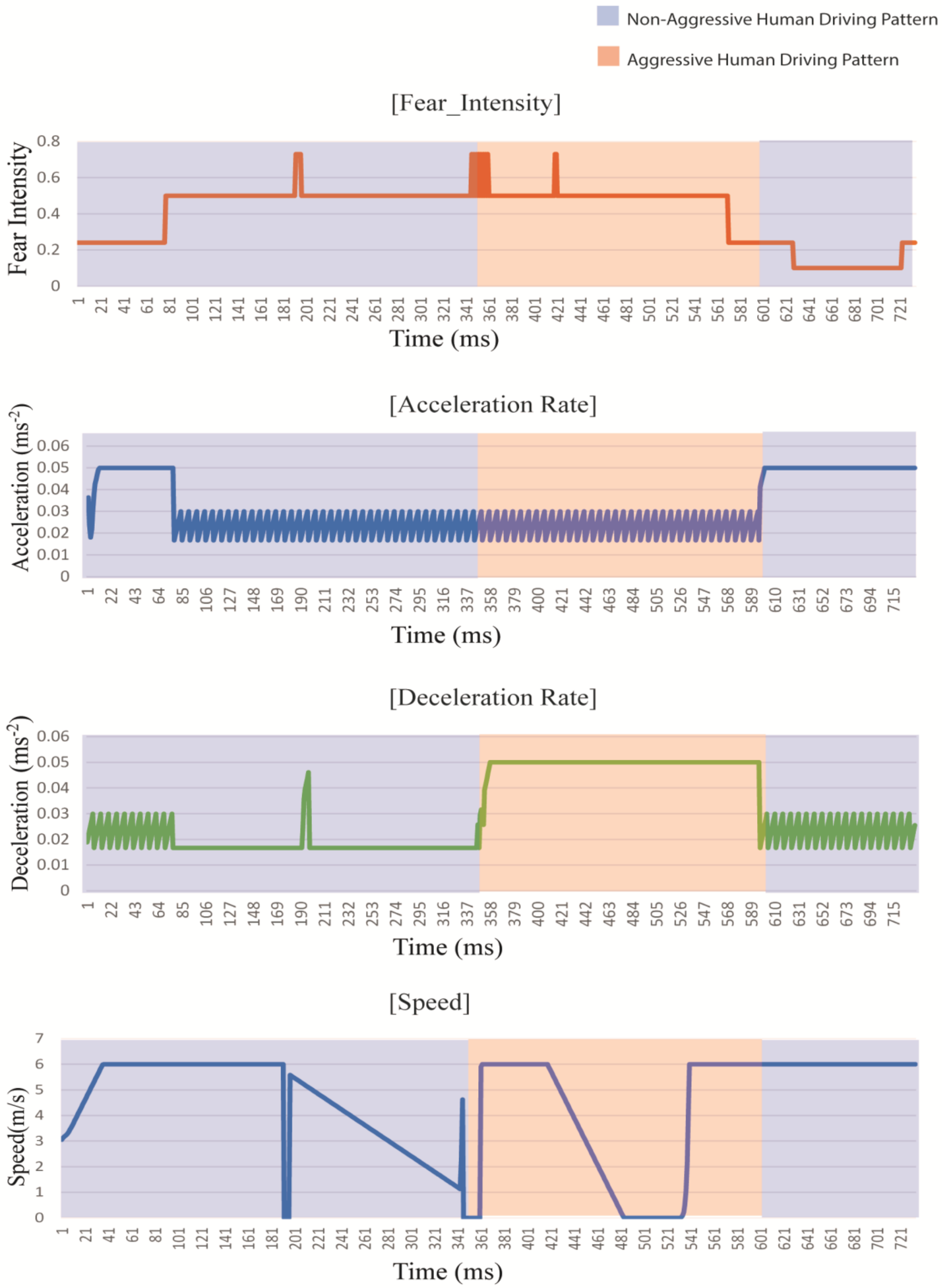

Figure. 16 Graphical depiction of results regarding performance measurement of EEEC_Agent based controller for rear end collision avoidance using AV-Human driven Car following test, when the leading human driver is in less aggressive mode or traffic pattern is less busy

no 2 and chose high deceleration and low acceleration rate. At 2041 ms the AV experienced the low fear intensity and it selects again the driving rule 1. Figure 16 presents the behaviour of prototype AV in another car following test, when the leading driver is in less aggressive or the traffic pattern is less bussy. Up to 342 ms, the AV experienced most of the time low and medium fear and there is only one switch between medium and high fear. From this fear intensity pattern, the AV learned that leading human driver is in non-aggressive mode and it selects driving rule 1 and 2 respectively. Between 342 ms to 422 ms, it experienced many switches between medium and high fear intensity over a short span of time and it learned that leading human driver has been shifted from non-aggressive mode to aggressive one or traffic pattern has become congested. In the result, it selects low acceleration and high deceleration rate to avoid the chances of collision. These results help us in validating the performance of proposed EEC_Agent along the proposed driving rules under the influence of fear emotion.

## C. Performance Measurement of EEC_Agent Based Controller For Tweaks Handling (AV-Pedestrian Collision Avoidance)

(1) The sudden appearance of a pedestrian in the front of prototype AV.

(2) The results of each test have been traced into a log file every millisecond and each test has been repeated 10 times.

**Results Discussion**- Figure 17 presents the performance validation of the behavior exhibited by EEC_Agent in the case of sudden appearance of a pedestrian. If we study the Fear Intenisty graph, then it can be seen that between 1ms to 378 ms the Fear level experienced by EEEC_Agent based controller fall in the range of very low fear because of very high distance from any obstacle ahead. Due to high distance and very low fear EEEC_Agent based controller select rule number 1 and start increasing its speed with high acceleration rate and keep maintained its maximum speed until the sudden appearance of pedestrian as shown in the speed graph. Then at 379 ms, the intensity level of fear suddenly jumped from very low fear to very high fear because of the occurrence of a tweak i.e. the sudden appearance of a pedestrian. After experiencing very high fear, EEEC_Agent based controller selects the rule number 3 and applied the brakes with no delay. We have tested the EEC_agent based controller for this scenario 10 times. Each time the results revealed that the proposed

EEEC_Agent based controller has the capability to tackle the highly unexpected road events and can avoid the collisions efficiently.

# 7 Qualitative Comparison With The Existing State of The Art

Brain emotional learning (BEL) has been utilised by Shahmirzadi and Langari in (17) to propose intelligent sliding mode control for rollover prevention in tractor-semitrailers. Though the authors have utilised the human brain-inspired emotion generation system, but have not utilised any authentic emotion computational mechanism, which helps in generating the emotions according to the rapid changes in the operating environment of vehicles. Comparatively, we have provided a rapid emotion learning model, which feel different levels of fear according to changes in the operating environment and help in developing an adaptive control mechanism to act rapidly and adaptively like humans and even better than them as the proposed emotions inspired controller do not stuck in dreading situations. Leu et al. (19) have modelled artificial driver with emotions and personality. The purpose of the research work was to study the behavioural aspects of human drivers and how their collective behaviour affects traffic performance. For this purpose, they have proposed a cognitive-affective inspired Driver mental model. The Driver mental model has been further equipped with a personality-Emotion model to represent different personalities of drivers using following emotions Anger, pleasure and sadness. These emotions have been generated using intensity-decay approach. Furthermore, the proposed driver model has been implemented using collision avoidance model in a customised traffic simulator. According to authors standard collision avoidance model takes less network transit time between pre-defined points as compared to the proposed cognitive-affective driver model because it considers drivers personalities and emotions in making decision making. Our research work also considers the role of human emotions in making driving decisions. However, our work is different in multi-aspects as compared to the Leu et al. (19) work. First of all, we have presented microscopic traffic model, instead of macroscopic traffic model as presented (19), which emulate the fear inspired tactical driving behaviour of individual driver to perform collision avoidance in road traffic. Second, we have utilised a well-defined appraisal model of OCC for emotion generation. The OCC model helps us to represent the true model of human driver behaviour during tactical level driving according to different road traffic events. Third, we have presented a proper agent-based model of the artificial driver and for this purpose, we have employed the exploratory agent-based modelling level of CABC framework. Fourth, we have properly defined the emotions eliciting conditions along with emotions intensity computation mechanism using fuzzy logic. Fifth, we have validated the emotion generation mechanism using real time scenarios. In comparison, the emotion generation process presented in (19) has not been validated, even not explained sufficiently. In (30), Zhenlog and Xiaoxia have presented emotion modelling of the human driver using fuzzy logic. For this purpose, they have considered Hidenori and Fukuda emotional space, which consists of four factors “Happy”, “Relief”, “Afraid”, and “Angry”. Authors have considered waiting time and road alignment as a stimulus to generate the different emotion levels. However, the paper has many limitations. The authors have not presented proper emotions eliciting conditions and intensity computation mechanism. Though, the fuzzy logic has been employed to model the emotions. But proper input/output fuzzy variables are not defined. Furthermore, the fuzzy rule base is not provided, which help in understanding the model and in making it generalised to be applicable to other road traffic problems, like collision detection and avoidance.

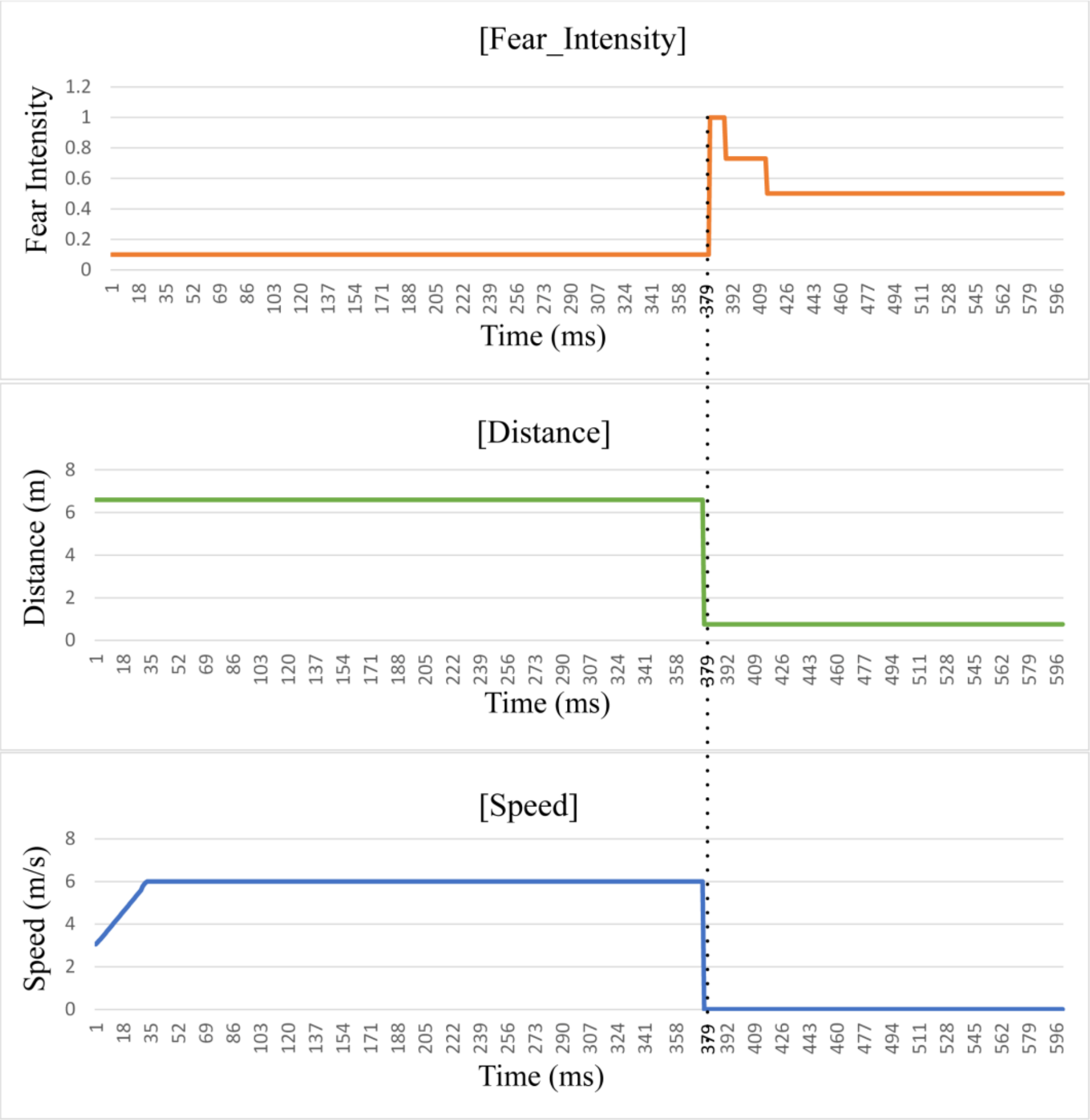

Figure. 17 Performance validation of the behavior exhibited by EEC_Agent in the case of sudden appearance of a pedestrian

According to Rizzi et al. (18), Fear learning is also an important ally for environmental adaptation as the brain constantly associates fear with newly experienced dangers. In (18), the authors have proposed situation appraisal mechanism for their robot. However, the authors have not utilised any well-investigated situation appraisal theory, which helps the robot to adapt the strategies according to the changing environment. Zoumpoulaki et al. (31) have proposed emotion and personality based BDI (Belief Desire Intention) framework to simulate the human decision making in natural disasters. Though the authors have discussed the effects of emotions and personality on the decisional and behavioural processes, they have not presented any details regarding emotions eliciting conditions and emotions intensity computation. Luo et al. (32) have presented an agent-based model to simulate the emotions inspired decision making of human beings using naturalistic decision-making theory named Recognition-Primed Decision (RPD). According to RPD theory, humans make decisions on the basis of their past experiences. Authors have reproduced the human-like behaviour in critical situations using this theory. However, this theory only focuses on the influences of emotions on decision making, whereas no emotion generation along emotion-eliciting conditions and emotion intensity computation mechanism has been presented. Belhaj et al. (33), have presented comprehensive emotions enabled agent model to reproduce the emotion generation process of human beings in emergency situations. For this purpose, they have employed complete OCC model.  The proposed agent model consists of perception, appraisal and behaviour modules. The authors have presented a proper emotion generation mechanism using event based emotions; react to the agent emotions and wellbeing emotions.  However, they have not presented emotions eliciting conditions along with emotions intensity computation mechanism. A tabular analysis regarding enhancements made in EEEC_Agent in comparison to the existing state of the art emotion enabled agents has been presented in table 16.

Table 16. Computation of Fear at high speed of Bullet AV with separation 17

| **References** | **Purpose** | **Human Brain Inspired Emotion Generation** | **Agent Based** | **Emotion Generation Mechanism** | **Emotion Elicitation Mechanism** | **Emotion Intensity Computation Mechanism** | **Validation of Emotion Generation Mechanism** |
|---|---|---|---|---|---|---|---|
| Shahmirzadi and Langari (17) | Brain emotional learning (BEL) based intelligent sliding mode control for rollover prevention in tractor-semitrailers | Yes | No | Mathematical Model | No | No | No |
| Rizzi et al. (18) | Fear based Collision avoidance mechanism for Robots | Yes | No | Mathematical Model | No | No | No |
| Leu et al. (19) | Modelling of artificial driver with emotions and personality | Yes | No | Intensity-decay approach | No | No | No |
| Zhenlog and Xiaoxia (30) | Emotion modelling of human driver | Yes | No | Hidenori and Fukuda emotional space | No | No | No |
| Zoumpoulaki et al. (31) | Simulation of the emotions inspired human decision mechanism | Yes | Yes | Self proposed non validated Appraisal theory | No | No | No |
| Luo et al. (32) | Agent based model to simulate the emotions inspired decision making of human beings | Yes | Yes | No | No | No | No |
| Belhaj et al. (33) | Agent model to reproduce the emotion generation process of human beings in emergency situations | Yes | Yes | Appraisal theory (OCC Model) | No | No | No |
| **Proposed System** | Agent model to reproduce the emotion generation process of human beings to propose the controller for rear end collision avoidance | Yes | Yes | Appraisal theory (OCC Model) | Yes | Yes | Yes |

In the existing literature, many researchers have developed rear end collision avoidance systems using different artificial intelligence algorithms. For example, Milanés et al. (9), have developed a fuzzy logic inspired rear end collision avoidance system for safe driving in congested traffic. For this purpose, two fuzzy logic based controllers have been utilized, which help them in collision warning and collision avoidance respectively. These controllers use Time to Collision (TTC) as a main consideration for collision warning and avoidance. This system has many drawbacks as compared to our proposed system. First of all, according to (8) Fuzzy logic based controllers are not suitable for real-time applications like collision avoidance because of the enormous number of fuzzy rules, which causes delay in decision making. Furthermore, the authors have not considered human emotions, which have actual impact on making collision avoidance decisions as human drivers accelerate/decelerate under the influence of different fear levels, which they feel according to different traffic situations. These acceleration/ deceleration maneuvers according to different fear levels help in avoiding the development of rear end collision situation with high comfort level of passengers. In another research work Genetic algorithm optimized fuzzy logic inspired rear end collision control has been proposed by Chen et al. (8) The authors have claimed that their work is better than the work of Milanés et al. (9), because they have considered the transient acceleration/deceleration of the trailing/leading cars along the TTC as well. Furthermore, they have utilized simple Genetic Algorithm (GA) as fuzzy rules optimization, which help them in producing fewer but the most effective rules. The authors have utilized genetic algorithm to optimize the fuzzy logic-based controller. They have claimed to optimize 39 rear end collision avoidance rules to 5 rules with the help of GA. However, in-depth analysis of their work reveals severe flaws in their work, which make their system impractical to build a real time efficient rear end collision avoidance system. The first main flaw is that the authors have a set number of generations as a stopping criterion, 30 in their case, for fuzzy rules optimization. After the evolution of 30 generations, GA provides a set of optimized fuzzy rules, which might be fully optimized or not because in GA the main stopping criteria is not the number of generations, but the convergence maturity. Hence the optimization of fuzzy rules in their research work is vague. In comparison to their so-called 5 rear end collision avoidance fuzzy rules, we have proposed three (3) fear inspired rules, which have proven to be more practical and efficient. The second flaw is that Chen et al. (8) claimed that they have proposed optimized fuzzy rules inspired by human driver behavior. However, they have not presented the threshold points which actually mimic the human driver acceleration/ deceleration decisions. This can be further explained as that they have constructed the fuzzy rules to mimic the human driver behavior to control the vehicles, but didn't provide the human drivers decision making, which is highly influenced by emotions and help the human drivers in defining the decision interrupts on the continuous timeline of the driving. To overcome this issue, we have utilized fear intensity as decision pivot, which help the AV to make more flexible decisions regarding acceleration, declaration and brake execution. The third drawback in the work of Chen et al. (8) is that they have just provided the inference rules; however, they have not modeled or defined the entity as an agent based system, which use these inference rules to act upon. We have utilized CABC framework to overcome this issue and have proposed a proper agent model, which emulate the main functions of the human driver like perception, fear generation, decision making and execution of rear end collision avoidance rules. Another problem with the research work of Chen et al. (8) is that their fuzzy rules only help in avoiding the collisions from vehicles, whereas other traffic cases like the sudden appearance of a pedestrian in the front of vehicle and static road hazards have not been considered. In comparison our proposed rear end collision avoidance system is very generic and can avoid collisions with leading vehicles as well as from the pedestrians and stationary road hazards. Chen et al. (8), have validated their simulation results using Arduino controller based toy vehicles. They have created a controlled environment for this purpose, ignoring the complexities of real road and real human driver behavior as a leading vehicle. They have utilized set of pre-fixed acceleration/deceleration rate of leading vehicle and programmed the following vehicle accordingly. In their validation test they considered a leading vehicle as AV. Whereas in contrast, we have validated our emotions inspired rear end collision avoidance controller on a real AV and tested it with real human-driven leading vehicle on a public road. Our proposed rear end collision avoidance controller has the capability to learn the traffic pattern or the behavior of the leading human driver and then it sets its acceleration and deceleration rate according to the behavior of the leading human driver. The results of our rigor validation reveal that using emotions inspired agent based rear end collision avoidance system, we can have an efficient alternative to human drivers by inhibiting their negative aspects and emulating their positive aspects. A tabular analysis of proposed EEEC_Agent based rear end collision avoidance controller with existing state of the art fuzzy logic based rear end collision avoidance controollers has been presented in table 17.

Table 17. Computation of Fear at high speed of Bullet AV with separation 17

| **State of the art** | **Purpose** | **Number of Rules** | **Human driver behaviour Inspired** | **Approach** | **Simulation** | **Practical Validation** |
|---|---|---|---|---|---|---|
| Milanés et al. (9) | Rear-end Collision avoidance | 39 | Yes | Fuzzy Logic | Yes | No |
| Chen et al. (8) | Rear-end Collision avoidance | 5 | Yes, claimed but ignoring the role of fear emotion in the rear end collision avoidance behaviour of human drivers | Fuzzy Logic | Yes | Toy-Based Car following test in a controlled environment |
| **Fear emotion inspired Proposed Controller** | Rear-end Collision avoidance | 3 | Yes, rules have been designed by keeping in mind the role of fear emotion in the rear end collision avoidance behaviour of human driver | Fear-Inspired | Yes | Real AV and real human driven vehicles based car following test in real road scenarios |

# 8 Conclusion

An enhanced version Emotion Enabled Cognitive Agent (EEEC_Agent) has been proposed to avoid the rear end collisions between autonomous vehicles. For this purpose, prospect based emotions defined by the OCC model are used to generate fear in EEC_Agent. SimConnector approach is used to join fuzzy logic environment results with NetLogo agent-based simulation. Furthermore, rigorous validation of EEEC_Agent functions like fear emotion generation mechanism, capability of learning leading human drivers' behavior, effectiveness of proposed rear end collision avoidance rules, and tweak handling has been performed using specially built prototype AV platform. In conclusion, it can be seen that the achieved results have proven that combining the human emotions with a cognitive agent, more robust type of collision avoidance system can be envisaged. Furthermore, our results are more practical and helpful for the automakers and academia for further research and enhancements in the rear end collision avoidance capabilities of AVs.

# References

1. Meng Q, Qu X. Estimation of rear-end vehicle crash frequencies in urban road tunnels. Accident Analysis & Prevention. 2012;48:254-63.
2. Harb R, Radwan E, Yan X, Abdel-Aty M. Light truck vehicles (LTVs) contribution to rear-end collisions. Accident Analysis & Prevention. 2007;39(5):1026-36.
3. Chen C, Zhang G, Tarefder R, Ma J, Wei H, Guan H. A multinomial logit model-Bayesian network hybrid approach for driver injury severity analyses in rear-end crashes. Accident Analysis & Prevention. 2015;80:76-88.
4. Nishimura N, Simms CK, Wood DP. Impact characteristics of a vehicle population in low speed front to rear collisions. Accident Analysis & Prevention. 2015;79:1-12.
5. Moon S, Moon I, Yi K. Design, tuning, and evaluation of a full-range adaptive cruise control system with collision avoidance. Control Engineering Practice. 2009;17(4):442-55.
6. Gracia L, Garelli F, Sala A. Reactive sliding-mode algorithm for collision avoidance in robotic systems. IEEE Transactions on Control Systems Technology. 2013;21(6):2391-9.
7. Van Den Berg J, Wilkie D, Guy SJ, Niethammer M, Manocha D, editors. LQG-obstacles: Feedback control with collision avoidance for mobile robots with motion and sensing uncertainty. Robotics and Automation (ICRA), 2012 IEEE International Conference on; 2012: IEEE.
8. Chen C, Li M, Sui J, Wei K, Pei Q. A genetic algorithm-optimized fuzzy logic controller to avoid rear-end collisions. Journal of Advanced Transportation. 2016.
9. Milanés V, Pérez J, Godoy J, Onieva E. A fuzzy aid rear-end collision warning/avoidance system. Expert Systems with Applications. 2012;39(10):9097-107.
10. Sato T, Akamatsu M. Applying Fuzzy Logic to Understanding Driving Modes in Real Road Environments. 2015.
11. Razzaq S, Riaz F, Mehmood T, Ratyal NI, editors. Multi-Factors Based Road Accident Prevention System. Computing, Electronic and Electrical Engineering (ICE Cube), 2016 International Conference on; 2016: IEEE.
12. Samiee S, Azadi S, Kazemi R, Eichberger A, Rogic B, Semmer M. Performance Evaluation of a Novel Vehicle Collision Avoidance Lane Change Algorithm. Advanced Microsystems for Automotive Applications 2015: Springer; 2016. p. 103-16.
13. Rumschlag G, Palumbo T, Martin A, Head D, George R, Commissaris RL. The effects of texting on driving performance in a driving simulator: The influence of driver age. Accident Analysis & Prevention. 2015;74:145-9.
14. Chan M, Singhal A. The emotional side of cognitive distraction: Implications for road safety. Accident Analysis & Prevention. 2013;50:147-54.
15. Lansdown TC, Stephens AN, Walker GH. Multiple driver distractions: A systemic transport problem. Accident Analysis & Prevention. 2015;74:360-7.
16. Zeng Y, Zhao F, Wang G, Zhang L, Xu B, editors. Brain-Inspired Obstacle Detection Based on the Biological Visual Pathway. International Conference on Brain and Health Informatics; 2016: Springer.
17. Shahmirzadi D, Langari R, Ricalde L, Sanchez E. Intelligent vs. sliding mode control in rollover prevention of tractor-semitrailers. International journal of vehicle autonomous systems. 2006;4(1):68-87.
18. Rizzi C, Johnson CG, Fabris F, Vargas PA. A Situation-Aware Fear Learning (SAFEL) model for robots. Neurocomputing. 2017;221:32-47.
19. Leu G, Curtis NJ, Abbass H. Modeling and evolving human behaviors and emotions in road traffic networks. Procedia-Social and Behavioral Sciences. 2012;54:999-1009.
20. Riaz F, Shah SI, Raees M, Shafi I, Iqbal A. Lateral Pre-crash sensing and avoidance in emotion enabled cognitive agent based vehicle-2-vehicle communication system. International Journal of Communication Networks and Information Security. 2013;5(2):127.
21. Ortony A, Clore GL, Collins A. The cognitive structure of emotions: Cambridge university press; 1990.
22. Niazi MA, Hussain A. Cognitive Agent-based Computing-I: A Unified Framework for Modeling Complex Adaptive Systems Using Agent-based & Complex Network-based Methods: Springer Science & Business Media; 2012.
23. Pradhan B. Brain, Mind, and Soul: Bridging the Gap. Yoga and Mindfulness Based Cognitive Therapy: Springer; 2015. p. 57-107.
24. Labouvie-Vief G. Emotions and Cognition: From Myth and Philosophy to Modern Psychology and Neuroscience. Integrating Emotions and Cognition Throughout the Lifespan: Springer; 2015. p. 1-16.
25. Li G, Amano T, Pare D, Nair SS. Impact of infralimbic inputs on intercalated amygdala neurons: a biophysical modeling study. Learning & Memory. 2011;18(4):226-40.
26. Zadeh LA. Fuzzy sets. Information and control. 1965;8(3):338-53.

27. El-Nasr MS, Yen J, editors. Agents, emotional intelligence and fuzzy logic. Fuzzy Information Processing Society-NAFIPS, 1998 Conference of the North American; 1998: IEEE.
28. Aly A, Tapus A. An online fuzzy-based approach for human emotions detection: An overview on the human cognitive model of understanding and generating multimodal actions. Intelligent Assistive Robots: Springer; 2015. p. 185-212.
29. Lotfi E, Akbarzadeh-T M-R. Adaptive brain emotional decayed learning for online prediction of geomagnetic activity indices. Neurocomputing. 2014;126:188-96.
30. Zhenlong L, Xiaoxia W, editors. Emotion modeling of the driver based on fuzzy logic. Intelligent Transportation Systems, 2009 ITSC'09 12th International IEEE Conference on; 2009: IEEE.
31. Zoumpoulaki A, Avradinis N, Vosinakis S, editors. A multi-agent simulation framework for emergency evacuations incorporating personality and emotions. Hellenic Conference on Artificial Intelligence; 2010: Springer.
32. Luo L, Zhou S, Cai W, Lees M, Low MYH, Sornum K. HumDPM: a decision process model for modeling human-like behaviors in time-critical and uncertain situations. Transactions on computational science XII: Springer; 2011. p. 206-30.
33. Belhaj M, Kebair F, Said LB, editors. Agent-based modeling and simulation of the emotional and behavioral dynamics of human civilians during emergency situations. German Conference on Multiagent System Technologies; 2014: Springer.